\documentclass{article}

\PassOptionsToPackage{numbers, compress}{natbib}
\usepackage[final]{neurips_2020}
\usepackage[utf8]{inputenc} 
\usepackage[T1]{fontenc}    
\usepackage{url}            
\usepackage{booktabs}       
\usepackage{amsfonts}       
\usepackage{nicefrac}       
\usepackage{microtype}      
\usepackage{bbold}
\usepackage{color}
\usepackage{bm}
\usepackage[table]{xcolor}
\definecolor{darkblue}{rgb}{0, 0.12, 0.55}
\definecolor{darkgreen}{rgb}{0, 0.55, 0.12}

\makeatletter
\renewcommand*{\@fnsymbol}[1]{\ensuremath{\ifcase#1\or
\dagger\or \ddagger\or
    \mathsection\or \mathparagraph\or \|\or **\or \dagger\dagger
    \or \ddagger\ddagger \else\@ctrerr\fi}}
\makeatother

\usepackage[
            bookmarksnumbered,
            bookmarksopen,
            colorlinks,
            linkcolor = darkblue,
            urlcolor  = darkblue,
            citecolor = darkblue,
            anchorcolor = darkblue
            ]{hyperref}

\usepackage{algorithmic}
\usepackage{algorithm}
\usepackage[algo2e,linesnumbered]{algorithm2e}
\usepackage{graphicx}
\usepackage{multirow}
\usepackage{epstopdf}
\usepackage{multicol}
\usepackage{tabularx}
\usepackage{arydshln}
\usepackage{multirow}
\usepackage{caption}
\usepackage{subfigure}
\usepackage{enumitem}
\usepackage{amsmath}
\usepackage{pifont}
\usepackage{amssymb}
\usepackage{lipsum}
\usepackage{pifont}
\usepackage{amsfonts}
\usepackage{mathrsfs}
\usepackage{amsthm}
\usepackage{float}
\usepackage{multirow}

\newcolumntype{L}[1]{>{\raggedright\let\newline\\\arraybackslash\hspace{0pt}}m{#1}}
\newcolumntype{Y}{>{\centering\arraybackslash}X}
\newcolumntype{s}{>{\hsize=.3\hsize}Y}
\newcolumntype{t}{>{\hsize=1.5\hsize}X}
\newcolumntype{b}{X}
\newcolumntype{u}{>{\hsize=0.8\hsize}Y}
\usepackage{makecell}
\usepackage{mathtools}

\title{Part-dependent Label Noise:\\ Towards Instance-dependent Label Noise}

\author{%
  Xiaobo Xia$^{1,2}$\quad
  Tongliang Liu$^{1}$\thanks{Correspondence to Tongliang Liu (tongliang.liu@sydney.edu.au).}\quad
  Bo Han$^{3}$\quad
  Nannan Wang$^2$\\
  \textbf{Mingming Gong$^4$\quad
  Haifeng Liu$^5$\quad
  Gang Niu$^6$\quad}
  \textbf{Dacheng Tao$^1$\quad
  Masashi Sugiyama$^{6,7}$}\\
  
  $^1$University of Sydney\quad
  $^2$Xidian University\\
  $^3$Hong Kong Baptist University\quad
  $^4$University of Melbourne\\
  $^5$Brain-Inspired Technology Co., Ltd\quad
  $^6$RIKEN\quad
  $^7$University of Tokyo
}
\begin{document}

\maketitle
\begin{abstract}
Learning with the \textit{instance-dependent} label noise is challenging, because it is hard to model such real-world noise. Note that there are psychological and physiological evidences showing that we humans perceive instances by decomposing them into parts. Annotators are therefore more likely to annotate instances based on the parts rather than the whole instances, where a wrong mapping from parts to classes may cause the instance-dependent label noise. Motivated by this human cognition, in this paper, we approximate the instance-dependent label noise by exploiting \textit{part-dependent} label noise. Specifically, since instances can be approximately reconstructed by a combination of parts, we approximate the instance-dependent \textit{transition matrix} for an instance by a combination of the transition matrices for the parts of the instance. The transition matrices for parts can be learned by exploiting anchor points (i.e., data points that belong to a specific class almost surely). Empirical evaluations on synthetic and real-world datasets demonstrate our method is superior to the state-of-the-art approaches for learning from the instance-dependent label noise.

\end{abstract}

\section{Introduction}

Learning with noisy labels can be dated back to \cite{angluin1988learning}, which has recently drawn a lot of attention, especially from the deep learning community, e.g.,  \cite{reed2014training,zhang2018generalized,kremer2018robust,goldberger2016training,patrini2017making,thekumparampil2018robustness,yu2018learning,liu2019peer,xu2019l_dmi,yu2019does,han2018co,malach2017decoupling,ren2018learning,jiang2018mentornet,ma2018dimensionality,tanaka2018joint,han2018masking,guo2018curriculumnet,veit2017learning,vahdat2017toward,li2017learning,li2019gradient,li2020dividemix,hu2020simple,lyu2020curriculum,nguyen2020self,yao2020dual,wu2020class2simi,yu2020label}. The main reason is that it is expensive and sometimes even infeasible to accurately label large-scale datasets \cite{karimi2019deep}; while it is relatively easy to obtain cheap but noisy datasets \cite{yu2018learning,vijayanarasimhan2014large,welinder2010online,yao2020searching,han2020sigua}.

Methods for dealing with label noise can be divided into two categories: model-free and model-based algorithms. In the first category, many heuristics reduce the side-effects of label noise without modeling it, e.g., extracting \textit{confident examples} with small losses \cite{han2018co,yu2019does,wang2019co}. Although these algorithms empirically work well, without modeling the label noise explicitly, their reliability cannot be guaranteed. For example, the small-loss-based methods rely on
accurate \textit{label noise rates}.

This inspires researchers to model and learn label noise \cite{goldberger2016training,scott2015rate,scott2013classification}. The \textit{transition matrix} $T(\bm{x})$ (i.e., a matrix-valued function) \cite{natarajan2013learning,cheng2017learning} was proposed to explicitly model the generation process of label noise, where $T_{ij}(\bm{x})=\text{Pr}(\bar{Y}=j|Y=i,X=\bm{x})$, $\text{Pr}(A)$ denotes as the probability of the event $A$, $X$ as the random variable for the instance, $\bar{Y}$ as the noisy label, and ${Y}$ as the latent clean label. Given the transition matrix, an optimal classifier defined by clean data can be learned by exploiting sufficient noisy data only \cite{patrini2017making,liu2016classification,yu2018learning}. The basic idea is that, the \textit{clean class posterior} can be inferred by using the \textit{noisy class posterior} (learned from the noisy data) and the transition matrix \cite{berthon2020idn}.

\begin{figure}[!tp]
\centering
\includegraphics[width=13.8cm,height=3.85cm]{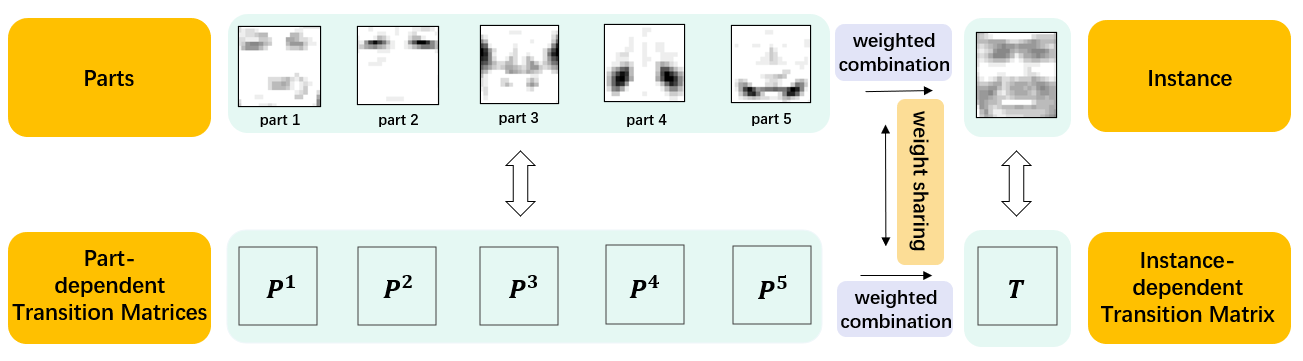}
\caption{The proposed method will learn the transition matrices for parts of instances. The instance-dependent transition matrix for each instance can be approximated by a weighted combination of the part-dependent transition matrices.}
\label{fig:flowchart}
\end{figure}

\vspace{-5pt}
However, in general, it is \textit{ill-posed} to learn the transition matrix $T(\bm{x})$ by only exploiting noisy data \cite{cheng2017learning,xia2019anchor}, i.e., the transition matrix is unidentifiable. Therefore, some assumptions are proposed to tackle this issue. For example, additional information is given \cite{berthon2020idn}; the matrix is symmetric \cite{menon2018learning}; the noise rates for instances are upper bounded \cite{cheng2017learning}, or even to be instance-independent 
\cite{xia2019anchor,han2018masking,patrini2017making,northcuttlearning,natarajan2013learning},  i.e., $\text{Pr}(\bar{Y}=j|Y=i,X=\bm{x})= \text{Pr}(\bar{Y}=j|Y=i)$. Note that there are specific applications where these assumptions are valid. That being said, in practice, these assumptions are hard to verify, and the gaps are large between instance-independent and instance-dependent transition matrices.

To handle the above problem, in this paper, we propose a new but practical assumption for instance-dependent label noise: \textit{The noise of an instance depends only on its parts}. We term this kind of noise as \textit{part-dependent} label noise. This assumption is motivated by that annotators usually annotate instances based on their parts rather than the whole instances. Specifically, there are psychological and physiological evidences showing that we humans perceive objects starting from their parts \cite{palmer1977hierarchical,wachsmuth1994recognition,logothetis1996visual}. There are also computational theories and learning algorithms showing that object recognition rely on parts-based representations \cite{biederman1987recognition,ullman1996high,dietterich1997solving,norouzi2013zero,hosseini2015deep,agarwal2004learning}. Since instances can be well reconstructed by combinations of parts \cite{lee1999learning,lee2001algorithms}, the part-dependence assumption should be mild in this sense. Intuitively, for a given instance, a combination of part-dependent transition matrices can well approximate the instance-dependent transition matrix, which will be empirically verified in Section \ref{sec:ablation}.

To fulfil the approximation, we need to learn the transition matrices for parts and the combination parameters. Since the parts are semantic \cite{lee1999learning}, their contributions to perceiving the instance could be similar in the contributions to understanding (or annotating) them \cite{biederman1987recognition,agarwal2004learning}. Therefore, it is natural to assume that for constructing the instance-dependent transition matrix, the combination parameters of  part-dependent transition matrices are identical to those of parts for reconstructing an instance. We illustrate this in Figure \ref{fig:flowchart}, where the combinations in the top and bottom panels share the same parameters. The transition matrices for parts can be learned by exploiting \textit{anchor points}, which are defined by instances that belong to a specific clean class with probability one \cite{liu2016classification}. Note that the assumption for combination parameters and the requirement of anchor points might be strong. If they are invalid, the part-dependent transition matrix might be poorly learned. To solve this issue, we also use the slack variable trick in \cite{xia2019anchor} to modify the instance-dependent transition matrix.

Extensive experiments on both synthetic and real-world label-noise datasets show that the part-dependent transition matrices can well address instance-dependent label noise. Specifically, when the instance-dependent label noise is heavy, i.e., 50\%, the proposed method outperforms state-of-the-art methods by almost $10\%$ of test accuracy on \textit{CIFAR-10}. More details can be found in Section \ref{section4}.

The rest of the paper is organized as follows. In Section \ref{section2}, we briefly review related work on modeling label noise and parts-based learning. In Section \ref{section3}, we discuss how to learn part-dependent transition matrices. In Section \ref{section4}, we provide empirical evaluations of our learning algorithm. In Section \ref{section5}, we conclude our paper. 

\vspace{-7pt}
\section{Related Work}\label{section2}
\textbf{Label noise models} \ \ Currently, there are three typical label noise models, i.e., the random classification noise (RCN) model \cite{biggio2011support,natarajan2013learning,manwani2013noise}, the class-conditional label noise (CCN) model \cite{patrini2017making,xia2019anchor,zhang2018generalized}, and the instance-dependent label noise (IDN) model \cite{berthon2020idn,cheng2017learning,du2015bcn}. Specifically, RCN assumes that clean labels flip randomly with a constant rate \cite{aslam1996complexity,angluin1988learning,kearns1993efficient}; CCN assumes that the flip rate depends on the latent clean class \cite{ma2018dimensionality,han2018co,yu2019does}; IDN considers the most general case of label noise, where the flip rate depends on its instance. However, IDN is non-identifiable without any additional assumption, which is hard to learn with only noisy data \cite{xia2019anchor}. The proposed part-dependent label noise (PDN) model assumes that the label noise depends on parts of instances, which could be an important ``intermediate'' model between CCN and IDN.

\textbf{Estimating the transition matrix} \ \ The transition matrix bridges the class posterior probabilities for noisy and clean data. It is essential to build \textit{classifier-/risk-consistent} estimators in label-noise learning \cite{patrini2017making, liu2016classification,scott2015rate, yu2018learning}. To estimate the transition matrix, a \textit{cross-validation} method is used for the binary classification task \cite{natarajan2013learning}. For the multi-classification task, the transition matrix could be learned by exploiting anchor points \cite{patrini2017making,yu2018efficient,yao2020towards}. To remove strong dependence on anchor points, data points having high noisy class posterior probabilities (similar to anchor points) can also be used to estimate the transition matrix via a slack variable trick \cite{xia2019anchor}. The slack variable is added to revise the transition matrix, which can be learned and validated together by using noisy data.

\textbf{Parts-based learning} \ \ Non-negative matrix factorization (NMF) \cite{dlee1999nmf} is the representative work of parts-based learning. It decomposes a non-negative data matrix into the product of two non-negative factor matrices. In contrast to principal components analysis (PCA) \cite{abdi2010pca} and vector quantization (VQ) \cite{gray1990Vector} that learn holistic but not parts-based representations, NMF allows additive but not subtractive combinations. Several variations extended the applicable range of NMF methods. For example, convex-NMF \cite{ding2008semi} restricts the basis vectors to be convex combinations of data. ONMF \cite{yoo2010nonnegative} imposes orthogonality constraints on data matrix, which achieves better performance than standard NMF in some applications. Semi-NMF \cite{trigeorgis2014deep} allows the data matrix and basis vectors to have mixed signs. LCNMF \cite{liu2017nmf} pushes the simplicial cone spanned by the bases to be large, and thus makes the learning algorithms robust. Truncated CauchyNMF \cite{guan2019nmf} can handle outliers by truncating large errors, which robustly learns the basis vectors on noisy datasets contaminated by outliers.  

\section{Part-dependent Label Noise}\label{section3}

\textbf{Preliminaries} \ \  Let $\bar{S}=\{(\bm{x}_i,\bar{y}_i)\}_{i=1}^{n}$ be the noisy training sample that contains instance-dependent label noise. Our aim is to learn a robust classifier from the noisy training sample that could assign clean labels for test data. In the rest of the paper, we use $A_{i\cdot}$ to denote the $i$-th row of the matrix $A$, $A_{\cdot j}$ the $j$-th column of the matrix $A$, and $A_{ij}$ the ${ij}$-th entry of the matrix $A$. We will use $\|\cdot\|_p$ as the $\ell_p$ norm of the matrices or vector, e.g., $\|A\|_p=\left(\sum_{ij}|A_{ij}|^p\right)^{1/p}$.
 
\textbf{Learning parts-based representations}\ \ 
NMF has been widely employed to learn parts-based representations \cite{dlee1999nmf}. Many variants of NMF were proposed to enlarge its application fields \cite{guan2019nmf,liu2017nmf,yoo2010nonnegative}, e.g., allowing the data matrix or/and the matrix of parts to have mixed signs \cite{ding2008semi}. For our problem, we do not require the matrix of parts to be non-negative, as our input data matrix is not restricted to be non-negative. However, we require the combination parameters (as known as new representation in the NMF community \cite{dlee1999nmf, liu2017nmf,guan2019nmf}) for each instance to be not only non-negative but also to have a unit $\ell_1$ norm. This is because we want to treat the parameters as the weights that measure how much the parts contribute to reconstructing the corresponding instance. 

Let $\bm{X}=[\bm{x}_1,\ldots, \bm{x}_n]\in \mathbb{R}^{d\times n}$ be the data matrix, where $d$ is the dimension of data points. The parts-based representation learning for the part-dependent label noise problem can be formulated as
\begin{equation}
\label{eq:optimize1}
	\begin{aligned}
        \min_{W\in\mathbb{R}^{d\times r}, \bm{h}(\bm{x}_i)\in\mathbb{R}^{r}_{+},\|\bm{h}(\bm{x}_i)\|_1=1, i=1,\ldots,n}& \quad  \sum\limits_{i=1}^n\|\bm{x}_i-W\bm{h}(\bm{x}_i)\|_2^2,  
    \end{aligned}
\end{equation}
where $W$ is the matrix of parts (each column of $W$ denotes a part of the instances) and the $\bm{h}(\bm{x}_i)$ denotes the combination parameters to reconstruct the instance $\bm{x}_i$. Eq.~\eqref{eq:optimize1} corresponds to the top panel of Figure \ref{fig:flowchart}, where parts are linearly combined to reconstruct the instance. Note that to fulfil the power of deep learning, the data matrix could consist of deep representations extracted by a deep neural network trained on the noisy training data. 

\textbf{Approximating instance-dependent transition matrices}\ \ Since there are computational theories \cite{biederman1987recognition,ullman1996high} and learning algorithms \cite{agarwal2004learning,hosseini2015deep} showing that object recognition rely on parts-based representations, it is therefore natural to model label noise on the part level. Thus, we propose a part-dependent noise (PDN) model, where label noise depends on parts rather than the whole instances. Specifically, for each part, e.g., $W_{\cdot j}$, we assume there is a 
part-dependent transition matrix, e.g., $P^j\in[0,1]^{c\times c}$. Since we have $r$ parts, there are $r$ different part-dependent transition matrices, i.e., $P^j, j=1,\ldots,r$. Similar to the idea that parts can be used to reconstruct instances, we exploit the idea that instance-dependent transition matrix can be approximated by a combination of part-dependent transition matrices, which is illustrated in the bottom panel of Figure \ref{fig:flowchart}. 

To approximate the instance-dependent transition matrices, we need to learn the part-dependent transition matrices and the combination parameters. However, they are not identifiable because it is ill-posed to factorize the instance-dependent transition matrix into the product of part-dependent transition matrices and combination parameters. Fortunately, we could identify the part-dependent transition matrices by assuming that \textit{the parameters for reconstructing the instance-dependent transition matrix are identical to those for reconstructing an instance}. The rational 
behind this assumption is that the learned parts are semantic \cite{lee1999learning}, and their contributions to perceiving the instance should be similar in the contributions to understanding and annotating them \cite{biederman1987recognition,agarwal2004learning}.
Let $\bm{h}(\bm{x})\in\mathbb{R}^r$ be the combination parameters to reconstruct the instance $\bm{x}$. The instance-dependent transition matrix $T(\bm{x})$ can be approximated by 
\begin{equation}
\label{eq:combine}
	T(\bm{x}) \approx \sum \limits_{j=1}^{r} \bm{h}_{j}(\bm{x})P^j.
\end{equation}
Note that $\bm{h}(\bm{x})$ can be learned via Eq.~\eqref{eq:optimize1}. The normalization constraint on the combination parameters, i.e., $\|\bm{h}(\bm{x})\|_1=1$, ensures that the combined matrix in the right-hand side of Eq.~\eqref{eq:combine} is also a valid transition matrix, which is non-negative and the sum of each row equals one.

\textbf{Learning the part-dependent transition matrices}\ \ Note that part-dependent transition matrices in Eq.~\eqref{eq:combine} are unknown. We will show that they can be learned by exploiting anchor points.
The concept of anchor points was proposed in \cite{liu2016classification}. They are defined in the clean data domain, i.e., an instance $\bm{x}^i$ is an anchor point of the $i$-th clean class if $\text{Pr}(Y = i|X=\bm{x}^i)$ is equal to one.

Let $\bm{x}^i$ be an anchor point of the $i$-th class. We have 
\begin{align}
\label{eq:anchor-estim}%
\text{Pr}(\bar{Y}=j|X=\bm{x}^i)=\sum_{k=1}^c\text{Pr}(\bar{Y}=j|Y=k,X=\bm{x}^i)\text{Pr}(Y=k|X=\bm{x}^i)=T_{ij}(\bm{x}^i),
\end{align}
where the first equation holds because of Law of total probability; the second equation holds because $\text{Pr}(Y=k|X=\bm{x}^i)=0$ for all $k\neq i$ and $\text{Pr}(Y=i|X=\bm{x}^i)=1$. As $[\text{Pr}(\bar{Y}=1|X=\bm{x}^i),\ldots,\text{Pr}(\bar{Y}=c|X=\bm{x}^i)]^\top$ can be unbiasedly learned \cite{bartlett2006convexity} by exploiting the noisy training sample and the anchor point $\bm{x}^i$, Eq.~\eqref{eq:anchor-estim} shows that the $i$-th row of the instance-dependent transition matrix $T(\bm{x}^i)$ can be unbiasedly learned. This sheds light on the learnability of the part-dependent transition matrices. Specifically, as shown in Figure \ref{fig:flowchart}, we are going to reconstruct the instance-dependent transition matrix by using a weighted combination of the part-dependent transition matrices. If the instance-dependent transition matrix\footnote[1]{Note that according to \eqref{eq:anchor-estim}, given an anchor point $\bm{x}_i$, the $i$-th row of its instance-dependent transition matrix can be learned and thus available.} and combination parameters are given, learning the part-dependent transition matrices is a convex problem.

Given an anchor point $\bm{x}^i$, we can learn the $i$-th rows of the part-dependent transition matrices by matching the $i$-th row of the reconstructed transition matrix, i.e., $\sum_{j=1}^r\bm{h}_{j}(\bm{x}^i)P_{i\cdot}^j$, with the $i$-th row of the instance-dependent transition matrix, i.e., $T_{i\cdot}(\bm{x}^i)$. Since we have $r$ part-dependent transition matrices, to identify all the entries of the $i$-th rows of the part-dependent transition matrices, we need at least $r$ 
anchor points of the $i$-th class to build $r$ equations. Let $(\bm{x}_1^i,\ldots, \bm{x}^i_k)$ be $k$ anchor points of the $i$-th class, where $k\geq r$. 
We robustly learn the $i$-th rows of the part-dependent transition matrices by minimizing the reconstruction error $\sum_{l=1}^k\|T_{i\cdot}(\bm{x}_l^i)-\sum_{j=1}^r\bm{h}_{j}(\bm{x}_l^i)P_{i\cdot}^j\|_2^2$ instead of solving $r$ equations. Therefore, we propose the following optimization problem to learn the part-dependent transition matrices:
\begin{equation}
\label{eq:optimize2}
	\begin{aligned}
        \min_{P^1,\ldots,P^r\in[0,1]^{c\times c}} &\quad \sum_{i=1}^c\sum_{l=1}^k\|T_{i\cdot}(\bm{x}_l^i)-\sum_{j=1}^r\bm{h}_{j}(\bm{x}_l^i)P_{i\cdot}^j\|_2^2,     \\
		&\quad \text{s.t.} \ \ \|P_{i\cdot}^j\|_1=1, i\in\{1,\ldots,c\}, j\in\{1,\ldots,r\},
\end{aligned}\end{equation}
where the sum over the index $i$ calculates the reconstruction error over all rows of transition matrices.
Note that in Eq.~\eqref{eq:optimize2}, we require that anchors for each class are given. 
If anchor points are not available, they can be learned from the noisy data as did in \cite{patrini2017making,liu2016classification,xia2019anchor}.

\begin{algorithm}[!tp]
 {\bfseries Input}: Noisy training sample $\mathcal{D}_\text{t}$, noisy validation data $\mathcal{D}_\text{v}$.

	1: Train a deep model by employing the noisy data $\mathcal{D}_\text{t}$ and $\mathcal{D}_\text{v}$;
	
	2: Get the deep representations of the instances by employing the trained deep network;
	
	3. Minimize Eq.~\eqref{eq:optimize1} to learn the parts and parameters;
	
	4: Learn the rows of instance-dependent transition matrices by anchor points according to Eq.~\eqref{eq:anchor-estim};
	
    5: Minimize Eq.~\eqref{eq:optimize2} to learn the part-dependent transition matrices;    
    
    6: Obtain the instance-dependent transition matrix for each instance according Eq.~\eqref{eq:combine};
	
{\bfseries Output}: $T(\bm{x})$. 
\caption{Part-dependent Matrices Learning Algorithm.}
\label{alg:mtml}
\end{algorithm}

\textbf{Implementation}\ \ The overall procedure to learn the part-dependent transition matrices is summarized in Algorithm \ref{alg:mtml}. Given only a noisy training sample set $\mathcal{D}_\text{t}$, we first learn deep representations of the instances. Note that we use a noisy validation set $\mathcal{D}_\text{v}$ to select the deep model. Then, we minimize Eq.~\eqref{eq:optimize1} to learn the combination parameters. The part-dependent transition matrices are learned by minimizing Eq.~\eqref{eq:optimize2}. Finally, we use the weighted combination to get an instance-dependent transition matrix for each instance according to Eq.~\eqref{eq:combine}. Note that as we learn the anchor points from the noisy training data, as did in \cite{patrini2017making,liu2016classification,xia2019anchor}, instances that are similar to anchor points will be learned if there are no anchor points available in the training data. Then, the instance-independent transition matrix will be poorly estimated. To address this issue, we employ the slack variable $\Delta T$ in \cite{xia2019anchor} to modify the instance-independent transition matrix.

\section{Experiments}\label{section4}
In this section, we first introduce the datasets, baselines, and implementation details used in the experiments (Section \ref{sec:4.1}). We next conduct an ablation study to show that the proposed method is not sensitive to the number of parts (Section \ref{sec:ablation}). Finally, we present and analyze the experimental results on synthetic and real-world noisy datasets to show the effectiveness of the proposed method (Section \ref{sec:4.3}).

\subsection{Experiment setup}\label{sec:4.1}

\textbf{Datasets} \ \ We verify the efficacy of our approach on the manually corrupted version of four datasets, i.e., \textit{F-MNIST} \cite{xiao2017fashion}, \textit{SVHN} \cite{netzer2011svhn}, \textit{CIFAR-10} \cite{krizhevsky2009learning}, \textit{NEWS} \cite{lang95}, and one real-world noisy dataset, i.e., \textit{Clothing1M} \cite{xiao2015learning}. \textit{F-MNIST}  contains 60,000 training images and 10,000 test images with 10 classes. \textit{SVHN} and \textit{CIFAR-10} both have 10 classes of images, but the former contains 73,257 training images and 26,032 test images, and the latter contains 50,000 training images and 10,000 test images. \textit{NEWS} contains 13,997 training texts and 6,000 test texts with 20 classes. We borrow the pre-trained word embeddings from GloVe \cite{pennington2014glove} for \textit{NEWS}. The four datasets contain clean data. We corrupted the training sets manually according to Algorithm \ref{alg:noise}. More details about this instance-dependent label noise generation approach can be found in Appendix B. IDN-$\tau$ means that the noise rate is controlled to be $\tau$.  All experiments on those datasets with synthetic instance-dependent label noise are repeated five times. \textit{Clothing1M} has 1M images with real-world noisy labels and 10k images with clean labels for testing. For all the datasets, we leave out 10\% of the noisy training examples as a noisy validation set, which is for model selection. We also conduct synthetic experiments on \textit{MNIST} \cite{LeCunmnist}. Due to the space limit, we put its corresponding experimental results in Appendix C. Significance tests are conducted to show whether experimental results are statistically significant. The details for significance tests can be found in Appendix D.

\textbf{Baselines and measurements} \ \ We compare the proposed method with the following state-of-the-art approaches: (i). CE, which trains the standard deep network with the cross entropy loss on noisy datasets. (ii). Decoupling \cite{malach2017decoupling}, which trains two networks on samples whose the predictions from the two networks are different. (iii). MentorNet \cite{jiang2018mentornet}, Co-teaching \cite{han2018co}, and Co-teaching+ \cite{yu2019does}. These approaches mainly handle noisy labels by training on instances with small loss values. (iv). Joint \cite{tanaka2018joint}, which jointly optimizes the sample labels and the network parameters. (v). DMI \cite{xu2019l_dmi}, which proposes a novel information-theoretic loss function for training deep neural networks robust to label noise. (vi). Forward \cite{patrini2017making}, Reweight \cite{liu2016classification}, and T-Revision \cite{xia2019anchor}. These approaches utilize a class-dependent transition matrix $T$ to correct the loss function. We use the classification accuracy to evaluate the performance of each model on the clean test set. Higher classification accuracy means that the algorithm is more robust to the label noise.

\begin{algorithm}[!tp]
 {\bfseries Input}: Clean samples $\{(\bm{x}_i, y_i)\}_{i=1}^{n}$; Noise rate $\tau$.
 
	1: Sample instance flip rates $q\in\mathbb{R}^{n}$ from the truncated normal distribution $\mathcal{N}(\tau,0.1^2,[0,1])$;
	
	2: Independently sample $w_1,w_2,\ldots,w_c$ from the standard normal distribution $\mathcal{N}(0,1^2)$;
	
	3: For $i=1,2,\ldots,n$ do
	
    4:\quad $p=\bm{x}_i\times w_{y_i}$;\hfill//generate instance-dependent flip rates
    
    5:\quad  $p_{y_i}=-\infty$;\hfill//control the diagonal entry of the instance-dependent transition matrix
    
    6:\quad  $p=q_i\times softmax(p)$;\hfill//make the sum of the off-diagonal entries of the $y_i$-th row to be $q_i$
    
    7:\quad  $p_{y_i}=1-q_i$;\hfill//set the diagonal entry to be 1-$q_i$
    
    8:\quad  Randomly choose a label from the label space according to the possibilities $p$ as noisy label $\bar{y}_i$;
    
	9: End for.
	
{\bfseries Output}: Noisy samples $\{(\bm{x}_i, \bar{y}_i)\}_{i=1}^{n}$
\caption{Instance-dependent Label Noise Generation}
\label{alg:noise}
\end{algorithm}

\textbf{Network structure and optimization} \ \ For fair comparison, all experiments are conducted on NVIDIA Tesla V100, and all methods are implemented by PyTorch. We use a ResNet-18 network for \textit{F-MNIST}, a ResNet-34 network for \textit{SVHN} and \textit{CIFAR-10}. We use a network with three convolutional layers and one fully connected layer for \textit{NEWS}. The transition matrix $T(\bm{x})$ for each instance $\bm{x}$ can be learned according to Algorithm \ref{alg:mtml}. Exploiting the transition matrices, we can bridge the class posterior probabilities for noisy and clean data. We first use SGD with momentum 0.9, weight decay $10^{-4}$, batch size 128, and an initial learning rate of $10^{-2}$ to initialize the network. The learning rate is divided by 10 at the 40th epochs and 80th epochs. We set 100 epochs in total. Then, the optimizer and learning rate are changed to Adam and $5 \times 10^{-7}$ to learn the classifier and slack variable. Note that the slack variable $\Delta T$ is initialized to be with all zero entries in the experiments. During the training, $T(\bm{x})+\Delta T$ can be ensured to be a valid transition matrix by first projecting their negative entries to be zero and then performing row normalization. Note that we do not use any data augmentation technique in the experiments. For \textit{Clothing1M}, we use a ResNet-50 pre-trained on ImageNet. Different from existing methods, we do not use the 50k clean training data or the 14k clean validation data but only exploit the 1M noisy data to learn the transition matrices and classifiers. 
Note that for real-world scenarios, it is more practical that no extra special clean data is provided to help adjust the model. After the  transition matrix $T(\bm{x})$ is obtained according to the Algorithm \ref{alg:mtml}, we use SGD with momentum 0.9, weight decay $10^{-3}$, batch size 32, and run with learning rate $10^{-3}$ for 10 epochs. For learning the classifier and the slack variable, Adam is used and learning rate is changed to $5 \times 10^{-7}$. Our implementation is available at \textcolor{darkblue}{https://github.com/xiaoboxia/Part-dependent-label-noise}.

\textbf{Explanation} \ \ We abbreviate our proposed method of learning with the \textit{\underline{p}ar\underline{t}-\underline{d}ependent} transition matrices as PTD. Methods with ``-F'' and ``-R'' mean that the instance-dependent transition matrices are exploited by using the Forward \cite{patrini2017making} method and the Reweight \cite{liu2016classification} method, respectively; Methods with ``-V'' means that the transition matrices are revised. Details for these methods can be found in Appendix A.

\begin{figure}[!t]
\centering
\subfigure[]{\label{fig2:a} 
\begin{minipage}[t]{0.5\linewidth}
\centering
\includegraphics[width=2.35in]{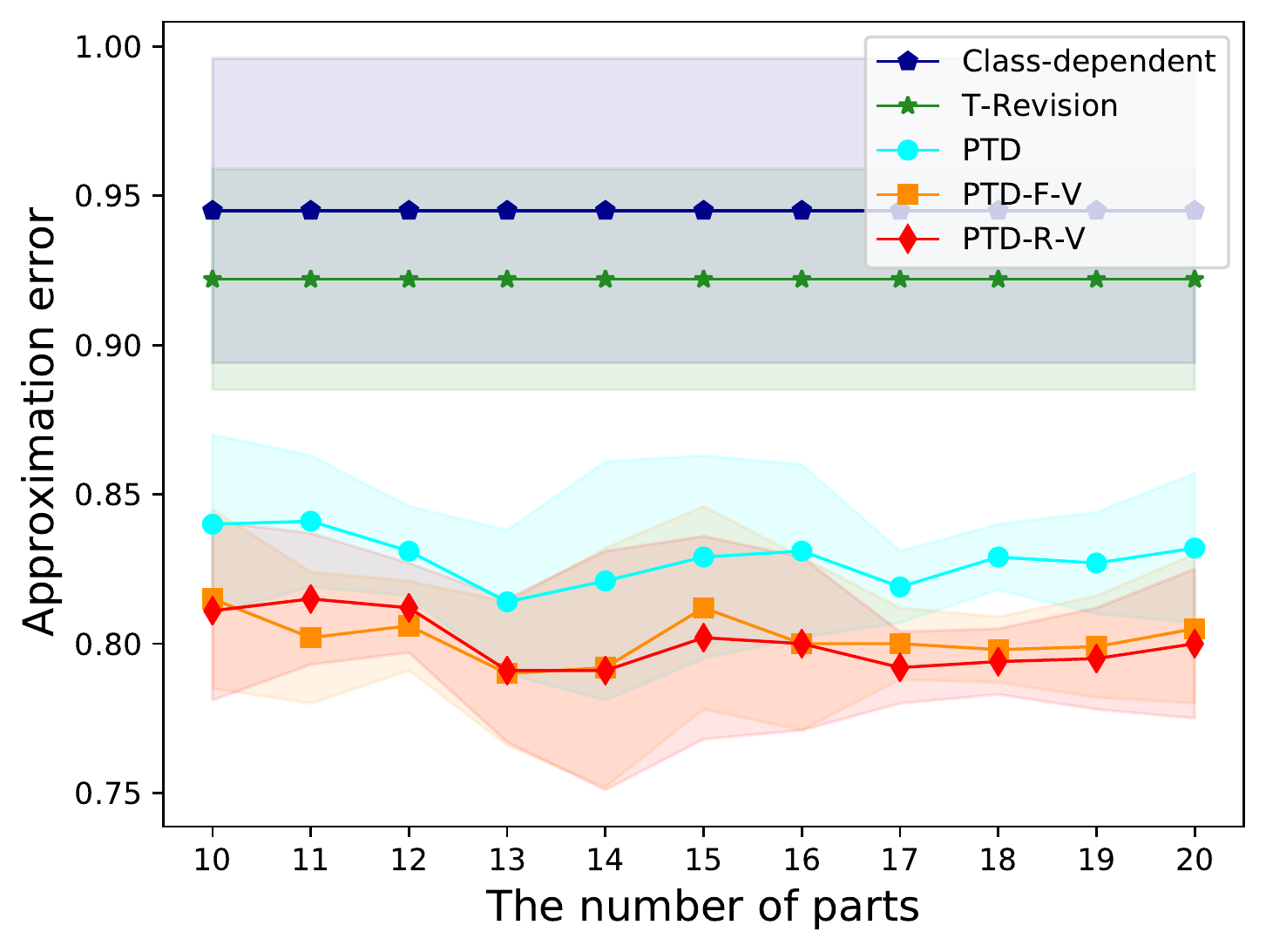}
\label{error}
\end{minipage}
}%
\subfigure[]{\label{fig2:b} 
\begin{minipage}[t]{0.5\linewidth}
\centering
\includegraphics[width=2.35in]{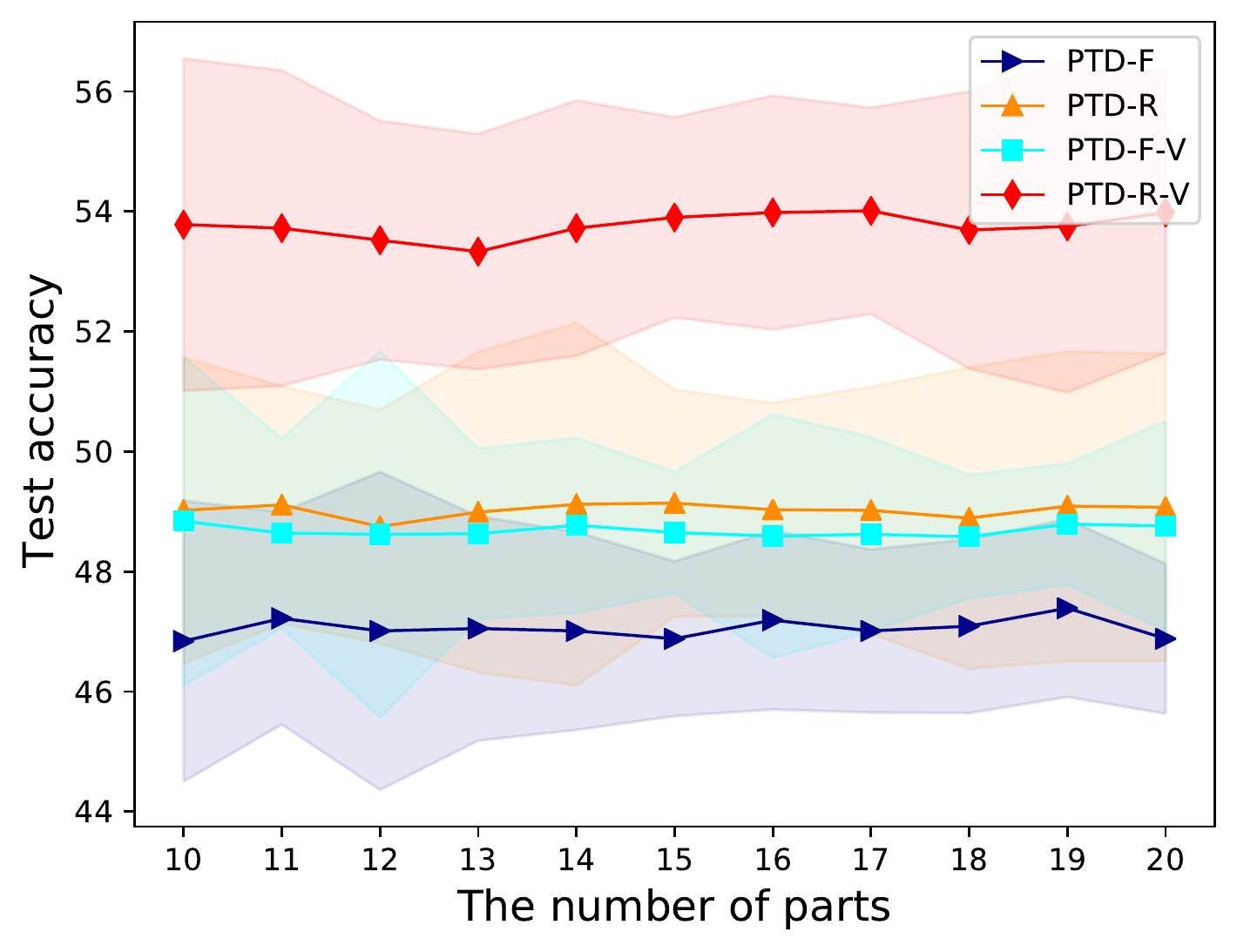}
\end{minipage}
}
\centering
\caption{Illustration of the transition matrix approximation error and the hyperparameter sensitivity. Figure (a) illustrates how the approximation error for the instance-dependent transition matrix varies by increasing the number of parts.  Figure (b) illustrates how the number of parts affects the test classification performance. The error bar for standard deviation in each figure has been shaded.}   
\label{fig2}
\end{figure}
\subsection{Ablation study}\label{sec:ablation}
 We have described that how to learn part-dependent transition matrices for approximating the instance-dependent transition matrix in Section \ref{section3}. To further prove that our proposed method is not sensitive to the number of parts, we perform ablation study in this subsection. The experiments are conducted on \textit{CIFAR-10} with 50\% noise rate. 
 
In Figure \ref{fig2:a}, we show how well the instance-dependent transition matrix can be approximated by employing the class-dependent transition matrix and the part-dependent transition matrix. We use $\ell_1$ norm to measure the difference. For each instance, we analyze the approximation error of 
a specific row rather than the whole transition matrix. The reason is that we only used one row of the instance-dependent transition matrix to generate the noisy label. Specifically, given an instance with clean class label $i$ (note that we have access to clean labels for the test data to conduct evaluation), we only exploit the $i$-th row of the instance-dependent transition matrix to flip the label from the class $i$ to another class. Note that  ``Class-dependent'' represents the standard class-dependent transition matrix learning methods \cite{liu2016classification,patrini2017making} and ``T-Revision'' represents the revision methods to learn class-dependent transition matrix \cite{xia2019anchor}. The Class-dependent and T-Revision methods are independent of parts. Their curves are therefore straight.
We can see that the part-dependent (PTD) transition matrix can achieve much smaller approximation error than the class-dependent (part-independent) transition matrix and the results are insensitive to the number of parts. 
Figure \ref{fig2:b} shows that the classification performance of our proposed method is robust and not sensitive to the change of the number of parts. More detailed experimental results can be found in Appendix E.

\begin{table}[!t]
	\centering
	\caption{Means and standard deviations (percentage) of classification accuracy on \textit{F-MNIST} with different label noise levels.}
	\label{tab:accu2}
	{
		\begin{tabular}{cccccc}
           	\toprule
			 ~& IDN-10\%      & IDN-20\%      & IDN-30\%    & IDN-40\%  & IDN-50\%         \\ \midrule
			 CE & 88.54$\pm$0.31 & 88.38$\pm$0.42 & 84.22$\pm$0.35 & 68.86$\pm$0.78 & 51.42$\pm$0.66\\
             Decoupling  & 89.27$\pm$0.31 & 86.50$\pm$0.35 & 85.33$\pm$0.47 & 78.54$\pm$0.53 & 57.32$\pm$2.11 \\ 
		     MentorNet & 90.00$\pm$0.34 & 87.02$\pm$0.41 & 86.02$\pm$0.82 & 80.12$\pm$0.76 & 58.62$\pm$1.36 \\
			 Co-teaching & 90.82$\pm$0.33 & 87.89$\pm$0.41 & 86.88$\pm$0.32  & 82.78$\pm$0.95 & 63.22$\pm$1.58 \\
			 Co-teaching+ & 90.92$\pm$0.51 & 89.77$\pm$0.45 & 88.52$\pm$0.45  & 83.57$\pm$1.77 & 59.32$\pm$2.77 \\
			 Joint & 70.24$\pm$0.99 & 56.83$\pm$0.45 & 51.27$\pm$0.67  & 44.24$\pm$0.78 & 30.45$\pm$0.45 \\
			 DMI & 91.98$\pm$0.62 & 90.33$\pm$0.21  & 84.81$\pm$0.44  & 69.01$\pm$1.87 & 51.64$\pm$1.78 \\
			 Forward & 89.05$\pm$0.43 & 88.61$\pm$0.43 & 84.27$\pm$0.46  & 70.25$\pm$1.28 & 57.33$\pm$3.75 \\
			 Reweight & 90.33$\pm$0.27 & 89.70$\pm$0.35 & 87.04$\pm$0.35  & 80.29$\pm$0.89 & 65.27$\pm$1.33 \\
			 T-Revision & 91.56$\pm$0.31 & 90.68$\pm$0.66 & 89.46$\pm$0.45  & 84.01$\pm$1.24 & 68.99$\pm$1.04 \\
			 \midrule
			 PTD-F & 90.48$\pm$0.17 & 90.01$\pm$0.31  & 87.42$\pm$0.65  & 83.89$\pm$0.49  & 68.25$\pm$2.61 \\
			 PTD-R & 91.01$\pm$0.22 & 90.03$\pm$0.32 & 87.68$\pm$0.42  & 84.03$\pm$0.52 & 72.43$\pm$1.76 \\
			 PTD-F-V & 91.61$\pm$0.19 & 90.79$\pm$0.29 & 89.33$\pm$0.33 & 85.32$\pm$0.36  & 71.89$\pm$2.54 \\
			 PTD-R-V & \textbf{92.01$\pm$0.35} & \textbf{91.08$\pm$0.46} & \textbf{89.66$\pm$0.43}  & \textbf{85.69$\pm$0.77} & \textbf{75.96$\pm$1.38} \\
             \bottomrule
		\end{tabular}
	}
\end{table}

\begin{table}[!t]
\vspace{-10pt}
	\centering
	\caption{Means and standard deviations (percentage) of classification accuracy on \textit{SVHN} with different instance-dependent label noise levels. }
	\label{tab:accu3}
	{
		\begin{tabular}{cccccc}
           	\toprule
			 ~& IDN-10\%      & IDN-20\%      & IDN-30\%    & IDN-40\%  & IDN-50\%         \\ \midrule
			 CE & 90.77$\pm$0.45 & 90.23$\pm$0.62 & 86.33$\pm$1.34 & 65.66$\pm$1.65 & 48.01$\pm$4.59\\

             Decoupling  & 90.49$\pm$0.15 & 90.47$\pm$0.66 & 85.27$\pm$0.34 & 82.57$\pm$1.45 & 42.56$\pm$2.79 \\ 
		     MentorNet & 90.28$\pm$0.12 & 90.37$\pm$0.37 & 86.49$\pm$0.49 & 83.75$\pm$0.75 & 40.27$\pm$3.14 \\
			 Co-teaching & 91.33$\pm$0.31 & 90.56$\pm$0.67 & 88.93$\pm$0.78  & 85.47$\pm$0.64 & 45.90$\pm$2.31 \\
			 Co-teaching+ & 93.05$\pm$1.20 & 91.05$\pm$0.82 & 85.33$\pm$2.71  & 57.24$\pm$3.77 & 42.56$\pm$3.65 \\
			 Joint & 86.01$\pm$0.34 & 78.58$\pm$0.72 & 76.34$\pm$0.56  & 65.14$\pm$1.72 & 46.78$\pm$3.77 \\
			 DMI & 93.51$\pm$1.09 & 93.22$\pm$0.62  & 91.78$\pm$1.54  & 69.34$\pm$2.45 & 48.93$\pm$2.34 \\
			 Forward & 90.89$\pm$0.63 & 90.65$\pm$0.27 & 87.32$\pm$0.59  & 78.46$\pm$2.58 & 46.27$\pm$3.90 \\
			 Reweight & 92.49$\pm$0.44 & 91.09$\pm$0.34 & 90.25$\pm$0.77  & 84.48$\pm$0.86 & 45.46$\pm$3.56 \\
			 T-Revision & 94.24$\pm$0.53 & 94.00$\pm$0.88 & 93.01$\pm$0.83  & 88.63$\pm$1.37 & 49.02$\pm$4.33 \\
			 \midrule
			 PTD-F & 93.62$\pm$0.61  & 92.77$\pm$0.45 & 90.11$\pm$0.94 & 87.25$\pm$0.77  & 54.82$\pm$4.65  \\
			 PTD-R & 93.21$\pm$0.45 & 92.36$\pm$0.68 & 90.57$\pm$0.42  & 86.78$\pm$0.63 & 55.88$\pm$3.73 \\
			 PTD-F-V & \textbf{94.70$\pm$0.37} & \textbf{94.39$\pm$0.62} & 92.07$\pm$0.59 & 90.56$\pm$1.21  & 57.92$\pm$4.32 \\
			 PTD-R-V & 94.44$\pm$0.37 & 94.23$\pm$0.46 & \textbf{93.11$\pm$0.78}  & \textbf{90.64$\pm$0.98} & \textbf{58.09$\pm$2.57} \\
             \bottomrule
		\end{tabular}
	}
\end{table}
\begin{table}[!t]
	\centering
	\caption{Means and standard deviations (percentage) of classification accuracy on \textit{CIFAR-10} with different label noise levels. }
	\label{tab:accu4}
	{
		\begin{tabular}{cccccc}
           	\toprule
			 ~& IDN-10\%      & IDN-20\%      & IDN-30\%    & IDN-40\%  & IDN-50\%         \\ \midrule
			 CE & 74.49$\pm$0.29 & 68.21$\pm$0.72 & 60.48$\pm$0.62 & 49.84$\pm$1.27 & 38.86$\pm$2.71\\
             Decoupling  & 74.09$\pm$0.78 & 70.01$\pm$0.66 & 63.05$\pm$0.65 & 44.27$\pm$1.91 & 38.63$\pm$2.32 \\ 
		     MentorNet & 74.45$\pm$0.66 & 70.56$\pm$0.34 & 65.42$\pm$0.79 & 46.22$\pm$0.98 & 39.89$\pm$2.62 \\
			 Co-teaching & 76.99$\pm$0.17 & 72.99$\pm$0.45 & 67.22$\pm$0.64  & 49.25$\pm$1.77 & 42.77$\pm$3.41 \\
			 Co-teaching+ & 74.27$\pm$1.20 & 71.07$\pm$0.77 & 64.77$\pm$0.58  & 47.73$\pm$2.32 & 39.47$\pm$2.14 \\
			 Joint & 76.89$\pm$0.37 & 73.89$\pm$0.34 & 69.03$\pm$0.79  & 54.75$\pm$5.98 & 44.72$\pm$7.72 \\
			 DMI & 75.02$\pm$0.45 & 69.89$\pm$0.33  & 61.88$\pm$0.64  & 51.23$\pm$1.18 & 41.45$\pm$1.97 \\
			 Forward & 73.45$\pm$0.23 & 68.99$\pm$0.62 & 60.21$\pm$0.75  & 47.17$\pm$2.96 & 40.75$\pm$2.09 \\
			 Reweight & 74.55$\pm$0.23 & 68.42$\pm$0.75 & 62.58$\pm$0.46  & 50.12$\pm$0.96 & 41.08$\pm$2.45 \\
			 T-Revision & 74.61$\pm$0.39 & 69.32$\pm$0.64 & 64.09$\pm$0.37  & 50.38$\pm$0.87 & 42.57$\pm$3.27 \\
			 \midrule
			 PTD-F & 76.01$\pm$0.45 & 73.45$\pm$0.62 & 65.25$\pm$0.84 & 49.88$\pm$0.85 & 46.88$\pm$1.25 \\
			 PTD-R & 78.71$\pm$0.22 & 75.02$\pm$0.73 & 71.86$\pm$0.42  & 56.15$\pm$0.45 & 49.07$\pm$2.56 \\
			 PTD-F-V & 76.29$\pm$0.38 & 73.88$\pm$0.61 & 69.01$\pm$0.47 & 50.43$\pm$0.62  & 48.76$\pm$2.01 \\
			 PTD-R-V & \textbf{79.01$\pm$0.20} & \textbf{76.05$\pm$0.53} & \textbf{72.28$\pm$0.49}  & \textbf{58.62$\pm$0.88} & \textbf{53.98$\pm$2.34} \\
             \bottomrule
		\end{tabular}
	}
\end{table}

\begin{table}[!t]
	\centering
	\caption{Means and standard deviations (percentage) of classification accuracy on \textit{NEWS} with different label noise levels. }
	\label{tab:accu6}
	{
		\begin{tabular}{cccccc}
           	\toprule
			 ~& IDN-10\%      & IDN-20\%      & IDN-30\%    & IDN-40\%  & IDN-50\%         \\ \midrule
			 CE & 69.58$\pm$0.42 & 66.80$\pm$0.36 & 63.11$\pm$0.74 & 58.37$\pm$0.88 & 54.75$\pm$1.62\\
             Decoupling  & 69.35$\pm$0.41 & 65.32$\pm$0.43 & 58.75$\pm$0.84 & 51.63$\pm$0.77 & 43.05$\pm$1.52 \\ 
		     MentorNet & 69.03$\pm$0.35 & 66.92$\pm$0.54 & 62.87$\pm$1.31 & 54.35$\pm$1.21 & 48.35$\pm$1.45 \\
			 Co-teaching & 69.37$\pm$0.29 & 67.99$\pm$0.76 & 64.15$\pm$0.89  & 56.36$\pm$0.71 & 52.32$\pm$1.03 \\
			 Co-teaching+ & 69.35$\pm$0.73 & 64.03$\pm$0.91 & 56.37$\pm$0.61  & 41.88$\pm$1.74 & 10.78$\pm$5.87 \\
			 Joint & 69.73$\pm$0.51 & 67.45$\pm$0.49 & 64.54$\pm$0.74  & 60.67$\pm$0.83 & 56.72$\pm$2.10 \\
			 DMI & 70.35$\pm$0.62 & 68.01$\pm$0.45 & 64.28$\pm$0.61  & 60.73$\pm$0.62 & 56.33$\pm$1.35 \\
			 Forward & 69.24$\pm$0.45 & 66.01$\pm$0.55 & 62.07$\pm$0.58  & 56.33$\pm$0.71 & 53.25$\pm$1.43 \\
			 Reweight & 70.25$\pm$0.30 & 68.42$\pm$0.77 & 65.05$\pm$0.93  & 59.37$\pm$1.32 & 57.31$\pm$3.51 \\
			 T-Revision & 70.72$\pm$0.32 & 69.91$\pm$0.49 & 67.28$\pm$0.81  & 61.78$\pm$0.99 & 59.29$\pm$2.07 \\
			 \midrule
			 PTD-F& 70.01$\pm$0.47 & 66.78$\pm$0.68 & 62.16$\pm$0.77  & 59.54$\pm$0.63 & 53.63$\pm$1.31 \\
			 PTD-R & 71.03$\pm$0.45 & 70.02$\pm$0.53 & 68.32$\pm$0.72  & 62.37$\pm$0.45 & 62.01$\pm$1.21 \\
			 PTD-F-V & 70.27$\pm$0.28 & 66.81$\pm$0.48 & 62.80$\pm$0.76  & 59.71$\pm$0.46 & 54.23$\pm$1.17 \\
			 PTD-R-V & \textbf{71.92$\pm$0.34} & \textbf{71.33$\pm$0.42} & \textbf{69.01$\pm$0.85}  & \textbf{63.17$\pm$0.58} & \textbf{62.77$\pm$0.98} \\
             \bottomrule
		\end{tabular}
	}
\end{table}

\subsection{Comparison with the State-of-the-Arts}\label{sec:4.3}
\textbf{Results on synthetic noisy datasets} \ \ Tables \ref{tab:accu2}, \ref{tab:accu3},  \ref{tab:accu4}, and \ref{tab:accu6} report the classification accuracy on the datasets of \textit{F-MNIST}, \textit{SVHN}, \textit{CIFAR-10}, and \textit{NEWS}, respectively. 

For \textit{F-MNIST} and \textit{SVHN}, in the easy cases, e.g., IDN-10\% and IDN-20\%, almost all methods work well. In the IDN-30\% case, the advantages of PTD begin to show. We surpassed all methods obviously except for T-Revision, e.g., the classification accuracy of PTD-R-V is 1.14\% higher than Co-teaching+ on \textit{F-MNIST}, 1.33\% higher than DMI on \textit{SVHN}. When the noise rate raises, T-Revision is gradually defeated. In the IDN-30\% case, the classification accuracy of PTD-R-V is 1.68\% and 2.01\% higher than T-Revision on \textit{SVHN} and \textit{CIFAR-10} respectively. Finally, in the hardest case, i.e., IDN-50\%, the superiority of PTD widens the gap of performance. The classification accuracy of PTD-R-V is 6.97\% and 9.07\% higher than the best baseline method.

For \textit{CIFAR-10}, the algorithms with the assist of PTD overtake the other methods with clear gaps. From IDN-10\% to IDN-50\% case, the advantages of our proposed method increase with the increasing of the noise rate. In the 10\% and 20\% cases, the performance of PTD-R-V is outstanding, i.e., the classification accuracy is 2.02\% and 2.16\% higher than the best baseline Joint. In the 30\% and 40\% case, the gap is expanded to 3.25\% and 3.87\%. Lastly, in the 50\% case, PTD-R-V outperforms state-of-the-art methods by almost 10\% of classification accuracy.

For \textit{NEWS}, the proposed method PTD-R-V consistently outperforms all the baseline methods. From IDN-10\% to IDN-40\% case, PTD-R-V clearly surpasses the best baseline T-Revision. In the hardest case, i.e., IDN-50\%, PTD-R-V outperforms all T-Revision by more than 3\% of test accuracy. 

To sum up, the synthetic experiments reveal that our method is powerful in handling \textit{instance-dependent} label noise particularly in the situation of high noise rates. 

\textbf{Results on real-world datasets} \ \ The proposed method outperforms the baselines as shown in Table~\ref{tab:accu5}, where the highest accuracy is bold faced. The comparison denotes that, the noise model of \textit{Clothing1M} dataset is more likely to be \textit{instance-dependent} noise, and our proposed method can better model instance-dependent noise than other methods.

\begin{table}[!t]
\vspace{-10pt}
	\centering
	\caption{Classification accuracy on \textit{Clothing1M}. In the experiments, only noisy samples are exploited to train and validate the deep model.}
	\label{tab:accu5}
	{
		\begin{tabular}{ccccccc}
           	\toprule
			 CE &Decoupling & MentorNet & Co-teaching & Co-teaching+ & Joint & DMI \\ \midrule
			 68.88 & 54.53 & 56.79 & 60.15 & 65.15 & 70.88 & 70.12 \\ \toprule
			 Forward & Reweight & T-Revision & PTD-F & PTD-R & PTD-F-V & PTD-R-V \\ \midrule
			 69.91 & 70.40 & 70.97 & 70.07 & 71.51 & 70.26 & \textbf{71.67}	\\ \toprule   
		\end{tabular}
	}
\end{table}

\section{Conclusion}\label{section5}
In this paper, we focus on learning with \textit{instance-dependent} label noise, which is a more general case of label noise but lacking understanding and learning. Inspired by parts-based learning, we exploit \textit{part-dependent} transition matrix to approximate \textit{instance-dependent} transition matrix, which is intuitive and learnable. Specifically, we first learn the parts of instances using all training examples. Then, we learn the part-dependent transition matrices by exploiting anchor points. Lastly, the instance-dependent transition matrix can be well approximated by a combination of the part-dependent transition matrices. Experimental results show our proposed method consistently outperforms existing methods, especially for the case of high-level noise rates. In future, we can extend the work in the following aspects. First, we can incorporate some prior knowledge of transition matrix and parts (e.g., sparsity), which improves parts-based learning. Second, we can introduce slack variables to modify the parameters for combination.

\section*{Broader Impact}
\textit{Instance-dependent} label noise is ubiquitous in the era of big data, which poses huge reliability threats for the traditional supervised learning algorithms. The instance-dependent label noise is more general and more realistic than \textit{instance-independent} label noise, but is hard to learn without any assumption. How to model such noise and reduce its side-effect should be considered by both research and industry communities. This research copes with \textit{instance-dependent} label noise based on the \textit{part-dependence} assumption. This assumption is milder and more practical. It is also supported by lots of evidences as stated in the paper. Outcomes of this research will promote the understanding of this kind of label noise and largely fills the gap between instance-independent and instance-dependent transition matrices. Open source algorithms and codes will benefit science, society, and the economy internationally through the applications to analyzing social, business, and health data.

The research may greatly benefit practitioners in industry communities, where large amounts of noisily labeled data are available. However, currently, the majority of machine learning applications are designed to fit high-quality labeled data. This research will improve tolerance for the errors of annotation and make cheap datasets with label noise be used effectively. However, inevitably, this research may have a negative impact on the jobs of annotators. 

The proposed method exploits the \textit{part-dependent} transition matrices to approximate the \textit{instance-dependent} transition matrix. If the \textit{part-dependent} transition matrices are poorly learned, the \textit{instance-dependent} transition matrix will be inaccurate. The classification performance of models therefore may be compromised. 

The proposed method does not leverage any bias in the data. 

\section*{Acknowledgments}
TLL was supported by Australian Research Council Project DE-190101473 and DP-180103424. BH was supported by the RGC Early Career Scheme No. 22200720, NSFC Young Scientists Fund No. 62006202, HKBU Tier-1 Start-up Grant, and HKBU CSD Start-up Grant. NNW was supported by National Natural Science Foundation of China under Grant 61922066 and Grant 61876142. DCT was supported by Project FL-170100117, DP-180103424, and IH-180100002. GN and MS were supported by JST AIP Acceleration Research Grant Number JPMJCR20U3, Japan. The authors would give special thanks to Pengqian Lu for helpful discussions and comments. The authors thank the reviewers and the meta-reviewer for their helpful and constructive comments on this work. 
\bibliographystyle{plainnat}
\bibliography{bib}

\begin{thebibliography}{81}
\providecommand{\natexlab}[1]{#1}
\providecommand{\url}[1]{\texttt{#1}}
\expandafter\ifx\csname urlstyle\endcsname\relax
  \providecommand{\doi}[1]{doi: #1}\else
  \providecommand{\doi}{doi: \begingroup \urlstyle{rm}\Url}\fi

\bibitem[A.Aslam and E.Decatur(1996)]{aslam1996complexity}
Javed A.Aslam and Scott E.Decatur.
\newblock On the sample complexity of noise- tolerant learning.
\newblock \emph{Information Processing Letters}, 1996.

\bibitem[Abdi and J.~Williams(2010)]{abdi2010pca}
Herv\'e Abdi and Lynne J.~Williams.
\newblock Principal component analysis.
\newblock \emph{wiley interdisciplinary reviews computational statistics},
  2\penalty0 (4):\penalty0 433--459, 2010.

\bibitem[Agarwal et~al.(2004)Agarwal, Awan, and Roth]{agarwal2004learning}
Shivani Agarwal, Aatif Awan, and Dan Roth.
\newblock Learning to detect objects in images via a sparse, part-based
  representation.
\newblock \emph{IEEE transactions on pattern analysis and machine
  intelligence}, 26\penalty0 (11):\penalty0 1475--1490, 2004.

\bibitem[Angluin and Laird(1988)]{angluin1988learning}
Dana Angluin and Philip Laird.
\newblock Learning from noisy examples.
\newblock \emph{Machine Learning}, 2\penalty0 (4):\penalty0 343--370, 1988.

\bibitem[Bartlett et~al.(2006)Bartlett, Jordan, and
  McAuliffe]{bartlett2006convexity}
Peter~L Bartlett, Michael~I Jordan, and Jon~D McAuliffe.
\newblock Convexity, classification, and risk bounds.
\newblock \emph{Journal of the American Statistical Association}, 101\penalty0
  (473):\penalty0 138--156, 2006.

\bibitem[Berthon et~al.(2020)Berthon, Han, Niu, Liu, and
  Sugiyama]{berthon2020idn}
Antonin Berthon, Bo~Han, Gang Niu, Tongliang Liu, and Masashi Sugiyama.
\newblock Confidence scores make instance-dependent label-noise learning
  possible.
\newblock \emph{arXiv preprint arXiv:2001.03772}, 2020.

\bibitem[Biederman(1987)]{biederman1987recognition}
Irving Biederman.
\newblock Recognition-by-components: a theory of human image understanding.
\newblock \emph{Psychological review}, 94\penalty0 (2):\penalty0 115, 1987.

\bibitem[Biggio et~al.(2011)Biggio, Nelson, and Laskov]{biggio2011support}
Battista Biggio, Blaine Nelson, and Pavel Laskov.
\newblock Support vector machines under adversarial label noise.
\newblock In \emph{ACML}, 2011.

\bibitem[Cheng et~al.(2020)Cheng, Liu, Ramamohanarao, and
  Tao]{cheng2017learning}
Jiacheng Cheng, Tongliang Liu, Kotagiri Ramamohanarao, and Dacheng Tao.
\newblock Learning with bounded instance-and label-dependent label noise.
\newblock In \emph{ICML}, 2020.

\bibitem[Dietterich et~al.(1997)Dietterich, Lathrop, and
  Lozano-P{\'e}rez]{dietterich1997solving}
Thomas~G Dietterich, Richard~H Lathrop, and Tom{\'a}s Lozano-P{\'e}rez.
\newblock Solving the multiple instance problem with axis-parallel rectangles.
\newblock \emph{Artificial intelligence}, 89\penalty0 (1-2):\penalty0 31--71,
  1997.

\bibitem[D.Lee and Seung(1999)]{dlee1999nmf}
Daniel D.Lee and H.Sebastian Seung.
\newblock Learning the parts of objects by non-negative matrix factorization.
\newblock \emph{Nature}, pages 788--791, 1999.

\bibitem[Du and Cai(2015)]{du2015bcn}
Jun Du and Zhihua Cai.
\newblock Modelling class noise with symmetric and asymmetric distributions.
\newblock In \emph{AAAI}, 2015.

\bibitem[Goldberger and Ben-Reuven(2017)]{goldberger2016training}
Jacob Goldberger and Ehud Ben-Reuven.
\newblock Training deep neural-networks using a noise adaptation layer.
\newblock In \emph{ICLR}, 2017.

\bibitem[Gray(1990)]{gray1990Vector}
Robert~M. Gray.
\newblock Vector quantization.
\newblock In \emph{Readings in Speech Recognition}, 1990.

\bibitem[Gretton et~al.(2009)Gretton, Smola, Huang, Schmittfull, Borgwardt, and
  Sch{\"o}lkopf]{gretton2009covariate}
Arthur Gretton, Alex Smola, Jiayuan Huang, Marcel Schmittfull, Karsten
  Borgwardt, and Bernhard Sch{\"o}lkopf.
\newblock Covariate shift by kernel mean matching.
\newblock \emph{Dataset shift in machine learning}, pages 131--160, 2009.

\bibitem[Guan et~al.(2019)Guan, Liu, Yangmuzi, Tao, and
  Steven~Davis]{guan2019nmf}
Naiyan Guan, Tongliang Liu, Zhang Yangmuzi, Dacheng Tao, and Larry
  Steven~Davis.
\newblock Truncated cauchy non-negative matrix factorization.
\newblock \emph{IEEE Transactions on pattern analysis and machine
  intelligence}, 41\penalty0 (1):\penalty0 246--259, 2019.

\bibitem[Guo et~al.(2018)Guo, Huang, Zhang, Zhuang, Dong, Scott, and
  Huang]{guo2018curriculumnet}
Sheng Guo, Weilin Huang, Haozhi Zhang, Chenfan Zhuang, Dengke Dong, Matthew~R
  Scott, and Dinglong Huang.
\newblock Curriculumnet: Weakly supervised learning from large-scale web
  images.
\newblock In \emph{ECCV}, pages 135--150, 2018.

\bibitem[Han et~al.(2018{\natexlab{a}})Han, Yao, Niu, Zhou, Tsang, Zhang, and
  Sugiyama]{han2018masking}
Bo~Han, Jiangchao Yao, Gang Niu, Mingyuan Zhou, Ivor Tsang, Ya~Zhang, and
  Masashi Sugiyama.
\newblock Masking: A new perspective of noisy supervision.
\newblock In \emph{NeurIPS}, pages 5836--5846, 2018{\natexlab{a}}.

\bibitem[Han et~al.(2018{\natexlab{b}})Han, Yao, Yu, Niu, Xu, Hu, Tsang, and
  Sugiyama]{han2018co}
Bo~Han, Quanming Yao, Xingrui Yu, Gang Niu, Miao Xu, Weihua Hu, Ivor Tsang, and
  Masashi Sugiyama.
\newblock Co-teaching: Robust training of deep neural networks with extremely
  noisy labels.
\newblock In \emph{NeurIPS}, pages 8527--8537, 2018{\natexlab{b}}.

\bibitem[Han et~al.(2020)Han, Niu, Yu, Yao, Xu, Tsang, and
  Sugiyama]{han2020sigua}
Bo~Han, Gang Niu, Xingrui Yu, Quanming Yao, Miao Xu, Ivor~W Tsang, and Masashi
  Sugiyama.
\newblock Sigua: Forgetting may make learning with noisy labels more robust.
\newblock In \emph{ICML}, 2020.

\bibitem[Hosseini-Asl et~al.(2015)Hosseini-Asl, Zurada, and
  Nasraoui]{hosseini2015deep}
Ehsan Hosseini-Asl, Jacek~M Zurada, and Olfa Nasraoui.
\newblock Deep learning of part-based representation of data using sparse
  autoencoders with nonnegativity constraints.
\newblock \emph{IEEE transactions on neural networks and learning systems},
  27\penalty0 (12):\penalty0 2486--2498, 2015.

\bibitem[Hu et~al.(2020)Hu, Li, and Yu]{hu2020simple}
Wei Hu, Zhiyuan Li, and Dingli Yu.
\newblock Simple and effective regularization methods for training on noisily
  labeled data with generalization guarantee.
\newblock In \emph{ICLR}, 2020.

\bibitem[Jiang et~al.(2018)Jiang, Zhou, Leung, Li, and
  Fei-Fei]{jiang2018mentornet}
Lu~Jiang, Zhengyuan Zhou, Thomas Leung, Li-Jia Li, and Li~Fei-Fei.
\newblock {MentorNet}: Learning data-driven curriculum for very deep neural
  networks on corrupted labels.
\newblock In \emph{ICML}, pages 2309--2318, 2018.

\bibitem[Karimi et~al.(2019)Karimi, Dou, Warfield, and
  Gholipour]{karimi2019deep}
Davood Karimi, Haoran Dou, Simon~K Warfield, and Ali Gholipour.
\newblock Deep learning with noisy labels: exploring techniques and remedies in
  medical image analysis.
\newblock \emph{arXiv preprint arXiv:1912.02911}, 2019.

\bibitem[Kearns(1993)]{kearns1993efficient}
Michael Kearns.
\newblock Efficient noise-tolerant learning from statistical queries.
\newblock In \emph{STOC}, 1993.

\bibitem[Kremer et~al.(2018)Kremer, Sha, and Igel]{kremer2018robust}
Jan Kremer, Fei Sha, and Christian Igel.
\newblock Robust active label correction.
\newblock In \emph{AISTATS}, pages 308--316, 2018.

\bibitem[Krizhevsky(2009)]{krizhevsky2009learning}
Alex Krizhevsky.
\newblock Learning multiple layers of features from tiny images.
\newblock Technical report, 2009.

\bibitem[Lang(1995)]{lang95}
Ken Lang.
\newblock Newsweeder: Learning to filter netnews.
\newblock In \emph{ICML}, pages 331--339, 1995.

\bibitem[LeCun et~al.()LeCun, Cortes, and Burges]{LeCunmnist}
Yann LeCun, Corinna Cortes, and Christopher~J.C. Burges.
\newblock The {MNIST} database of handwritten digits.
\newblock \emph{http://yann.lecun.com/exdb/mnist/}.

\bibitem[Lee and Seung(1999)]{lee1999learning}
Daniel~D Lee and H~Sebastian Seung.
\newblock Learning the parts of objects by non-negative matrix factorization.
\newblock \emph{Nature}, 401\penalty0 (6755):\penalty0 788--791, 1999.

\bibitem[Lee and Seung(2001)]{lee2001algorithms}
Daniel~D Lee and H~Sebastian Seung.
\newblock Algorithms for non-negative matrix factorization.
\newblock In \emph{NeurIPS}, pages 556--562, 2001.

\bibitem[Li et~al.(2020{\natexlab{a}})Li, Socher, and Hoi]{li2020dividemix}
Junnan Li, Richard Socher, and Steven~C.H. Hoi.
\newblock Dividemix: Learning with noisy labels as semi-supervised learning.
\newblock In \emph{ICLR}, 2020{\natexlab{a}}.

\bibitem[Li et~al.(2020{\natexlab{b}})Li, Soltanolkotabi, and
  Oymak]{li2019gradient}
Mingchen Li, Mahdi Soltanolkotabi, and Samet Oymak.
\newblock Gradient descent with early stopping is provably robust to label
  noise for overparameterized neural networks.
\newblock In \emph{AISTATS}, 2020{\natexlab{b}}.

\bibitem[Li et~al.(2017)Li, Yang, Song, Cao, Luo, and Li]{li2017learning}
Yuncheng Li, Jianchao Yang, Yale Song, Liangliang Cao, Jiebo Luo, and Li-Jia
  Li.
\newblock Learning from noisy labels with distillation.
\newblock In \emph{ICCV}, pages 1910--1918, 2017.

\bibitem[Liu et~al.(2010)Liu, Li, and I.~Jordan]{ding2008semi}
Ding Liu, Chris~H.Q., Tao Li, and Michael I.~Jordan.
\newblock Convex and semi-nonnegative matrix factorizations.
\newblock \emph{IEEE Transactions on pattern analysis and machine
  intelligence}, 32\penalty0 (1):\penalty0 45--55, 2010.

\bibitem[Liu and Tao(2016)]{liu2016classification}
Tongliang Liu and Dacheng Tao.
\newblock Classification with noisy labels by importance reweighting.
\newblock \emph{IEEE Transactions on pattern analysis and machine
  intelligence}, 38\penalty0 (3):\penalty0 447--461, 2016.

\bibitem[Liu et~al.(2017)Liu, Gong, and Tao]{liu2017nmf}
Tongliang Liu, Mingming Gong, and Dacheng Tao.
\newblock Large cone non-negative matrix factorization.
\newblock \emph{IEEE Transactions on Neural Networks and Learning Systems},
  28\penalty0 (9):\penalty0 2129--2141, 2017.

\bibitem[Liu and Guo(2020)]{liu2019peer}
Yang Liu and Hongyi Guo.
\newblock Peer loss functions: Learning from noisy labels without knowing noise
  rates.
\newblock In \emph{ICML}, 2020.

\bibitem[Logothetis and Sheinberg(1996)]{logothetis1996visual}
Nikos~K Logothetis and David~L Sheinberg.
\newblock Visual object recognition.
\newblock \emph{Annual review of neuroscience}, 19\penalty0 (1):\penalty0
  577--621, 1996.

\bibitem[Lyu and Tsang(2020)]{lyu2020curriculum}
Yueming Lyu and Ivor~W. Tsang.
\newblock Curriculum loss: Robust learning and generalization against label
  corruption.
\newblock In \emph{ICLR}, 2020.

\bibitem[Ma et~al.(2018)Ma, Wang, Houle, Zhou, Erfani, Xia, Wijewickrema, and
  Bailey]{ma2018dimensionality}
Xingjun Ma, Yisen Wang, Michael~E Houle, Shuo Zhou, Sarah~M Erfani, Shu-Tao
  Xia, Sudanthi Wijewickrema, and James Bailey.
\newblock Dimensionality-driven learning with noisy labels.
\newblock In \emph{ICML}, pages 3361--3370, 2018.

\bibitem[Malach and Shalev-Shwartz(2017)]{malach2017decoupling}
Eran Malach and Shai Shalev-Shwartz.
\newblock Decoupling" when to update" from" how to update".
\newblock In \emph{NeurIPS}, pages 960--970, 2017.

\bibitem[Manwani and Sastry(2013)]{manwani2013noise}
Naresh Manwani and P.S. Sastry.
\newblock Noise tolerance under risk minimization.
\newblock \emph{IEEE Transactions on Cybernetics}, 2013.

\bibitem[Menon et~al.(2018)Menon, Van~Rooyen, and Natarajan]{menon2018learning}
Aditya~Krishna Menon, Brendan Van~Rooyen, and Nagarajan Natarajan.
\newblock Learning from binary labels with instance-dependent noise.
\newblock \emph{Machine Learning}, 107\penalty0 (8-10):\penalty0 1561--1595,
  2018.

\bibitem[Natarajan et~al.(2013)Natarajan, Dhillon, Ravikumar, and
  Tewari]{natarajan2013learning}
Nagarajan Natarajan, Inderjit~S Dhillon, Pradeep~K Ravikumar, and Ambuj Tewari.
\newblock Learning with noisy labels.
\newblock In \emph{NeurIPS}, pages 1196--1204, 2013.

\bibitem[Netzer et~al.(2011)Netzer, Wang, Coates, Bissacco, Wu, and
  Y.Ng]{netzer2011svhn}
Yuval Netzer, Tao Wang, Adam Coates, Alessandro Bissacco, Bo~Wu, and Andrew
  Y.Ng.
\newblock Reading digits in natural images with unsupervised feature learning.
\newblock In \emph{NIPS Workshop on Deep Learning and Unsupervised Feature
  Learning}, 2011.

\bibitem[Nguyen et~al.(2020)Nguyen, Mummadi, Ngo, Nguyen, Beggel, and
  Brox]{nguyen2020self}
Duc~Tam Nguyen, Chaithanya~Kumar Mummadi, Thi Phuong~Nhung Ngo, Thi Hoai~Phuong
  Nguyen, Laura Beggel, and Thomas Brox.
\newblock Self: Learning to filter noisy labels with self-ensembling.
\newblock In \emph{ICLR}, 2020.

\bibitem[Norouzi et~al.(2013)Norouzi, Mikolov, Bengio, Singer, Shlens, Frome,
  Corrado, and Dean]{norouzi2013zero}
Mohammad Norouzi, Tomas Mikolov, Samy Bengio, Yoram Singer, Jonathon Shlens,
  Andrea Frome, Greg~S Corrado, and Jeffrey Dean.
\newblock Zero-shot learning by convex combination of semantic embeddings.
\newblock In \emph{NeurIPS}, 2013.

\bibitem[Northcutt et~al.(2017)Northcutt, Wu, and Chuang]{northcuttlearning}
Curtis~G Northcutt, Tailin Wu, and Isaac~L Chuang.
\newblock Learning with confident examples: Rank pruning for robust
  classification with noisy labels.
\newblock In \emph{UAI}, 2017.

\bibitem[Palmer(1977)]{palmer1977hierarchical}
Stephen~E Palmer.
\newblock Hierarchical structure in perceptual representation.
\newblock \emph{Cognitive psychology}, 9\penalty0 (4):\penalty0 441--474, 1977.

\bibitem[Patrini et~al.(2017)Patrini, Rozza, Krishna~Menon, Nock, and
  Qu]{patrini2017making}
Giorgio Patrini, Alessandro Rozza, Aditya Krishna~Menon, Richard Nock, and
  Lizhen Qu.
\newblock Making deep neural networks robust to label noise: A loss correction
  approach.
\newblock In \emph{CVPR}, pages 1944--1952, 2017.

\bibitem[Pennington et~al.(2014)Pennington, Socher, and
  Manning]{pennington2014glove}
Jeffrey Pennington, Richard Socher, and Christopher~D. Manning.
\newblock Glove: Global vectors for word representation.
\newblock In \emph{EMNLP}, pages 1532--1543, 2014.

\bibitem[Reed et~al.(2015)Reed, Lee, Anguelov, Szegedy, Erhan, and
  Rabinovich]{reed2014training}
Scott~E Reed, Honglak Lee, Dragomir Anguelov, Christian Szegedy, Dumitru Erhan,
  and Andrew Rabinovich.
\newblock Training deep neural networks on noisy labels with bootstrapping.
\newblock In \emph{ICLR}, 2015.

\bibitem[Ren et~al.(2018)Ren, Zeng, Yang, and Urtasun]{ren2018learning}
Mengye Ren, Wenyuan Zeng, Bin Yang, and Raquel Urtasun.
\newblock Learning to reweight examples for robust deep learning.
\newblock In \emph{ICML}, pages 4331--4340, 2018.

\bibitem[Scott(2015)]{scott2015rate}
Clayton Scott.
\newblock A rate of convergence for mixture proportion estimation, with
  application to learning from noisy labels.
\newblock In \emph{AISTATS}, pages 838--846, 2015.

\bibitem[Scott et~al.(2013)Scott, Blanchard, and
  Handy]{scott2013classification}
Clayton Scott, Gilles Blanchard, and Gregory Handy.
\newblock Classification with asymmetric label noise: Consistency and maximal
  denoising.
\newblock In \emph{COLT}, pages 489--511, 2013.

\bibitem[Sedgwick(2010)]{sedgwick2010independent}
Philip Sedgwick.
\newblock Independent samples t test.
\newblock \emph{BMJ}, 340:\penalty0 c2673, 2010.

\bibitem[Tanaka et~al.(2018)Tanaka, Ikami, Yamasaki, and
  Aizawa]{tanaka2018joint}
Daiki Tanaka, Daiki Ikami, Toshihiko Yamasaki, and Kiyoharu Aizawa.
\newblock Joint optimization framework for learning with noisy labels.
\newblock In \emph{CVPR}, 2018.

\bibitem[Thekumparampil et~al.(2018)Thekumparampil, Khetan, Lin, and
  Oh]{thekumparampil2018robustness}
Kiran~K Thekumparampil, Ashish Khetan, Zinan Lin, and Sewoong Oh.
\newblock Robustness of conditional gans to noisy labels.
\newblock In \emph{NeurIPS}, pages 10271--10282, 2018.

\bibitem[Trigeorgis et~al.(2014)Trigeorgis, Bousmalis, Zafeiriou, and
  Schuller]{trigeorgis2014deep}
George Trigeorgis, Konstantinos Bousmalis, Stefanos Zafeiriou, and Bjoern
  Schuller.
\newblock A deep semi-nmf model for learning hidden representations.
\newblock In \emph{ICML}, pages 1692--1700, 2014.

\bibitem[Ullman et~al.(1996)]{ullman1996high}
Shimon Ullman et~al.
\newblock \emph{High-level vision: Object recognition and visual cognition},
  volume~2.
\newblock MIT press Cambridge, MA, 1996.

\bibitem[Vahdat(2017)]{vahdat2017toward}
Arash Vahdat.
\newblock Toward robustness against label noise in training deep discriminative
  neural networks.
\newblock In \emph{NeurIPS}, pages 5596--5605, 2017.

\bibitem[Veit et~al.(2017)Veit, Alldrin, Chechik, Krasin, Gupta, and
  Belongie]{veit2017learning}
Andreas Veit, Neil Alldrin, Gal Chechik, Ivan Krasin, Abhinav Gupta, and Serge
  Belongie.
\newblock Learning from noisy large-scale datasets with minimal supervision.
\newblock In \emph{CVPR}, pages 839--847, 2017.

\bibitem[Vijayanarasimhan and Grauman(2014)]{vijayanarasimhan2014large}
Sudheendra Vijayanarasimhan and Kristen Grauman.
\newblock Large-scale live active learning: Training object detectors with
  crawled data and crowds.
\newblock \emph{International journal of computer vision}, 108\penalty0
  (1-2):\penalty0 97--114, 2014.

\bibitem[Wachsmuth et~al.(1994)Wachsmuth, Oram, and
  Perrett]{wachsmuth1994recognition}
E~Wachsmuth, MW~Oram, and DI~Perrett.
\newblock Recognition of objects and their component parts: responses of single
  units in the temporal cortex of the macaque.
\newblock \emph{Cerebral Cortex}, 4\penalty0 (5):\penalty0 509--522, 1994.

\bibitem[Wang et~al.(2019)Wang, Wang, Wang, Shi, and Mei]{wang2019co}
Xiaobo Wang, Shuo Wang, Jun Wang, Hailin Shi, and Tao Mei.
\newblock Co-mining: Deep face recognition with noisy labels.
\newblock In \emph{ICCV}, pages 9358--9367, 2019.

\bibitem[Welinder and Perona(2010)]{welinder2010online}
Peter Welinder and Pietro Perona.
\newblock Online crowdsourcing: rating annotators and obtaining cost-effective
  labels.
\newblock In \emph{CVPR-Workshop}, pages 25--32, 2010.

\bibitem[Wu et~al.(2020)Wu, Xia, Liu, Han, Gong, Wang, Liu, and
  Niu]{wu2020class2simi}
Songhua Wu, Xiaobo Xia, Tongliang Liu, Bo~Han, Mingming Gong, Nannan Wang,
  Haifeng Liu, and Gang Niu.
\newblock Class2simi: A new perspective on learning with label noise.
\newblock \emph{arXiv preprint arXiv:2006.07831}, 2020.

\bibitem[Xia et~al.(2019)Xia, Liu, Wang, Han, Gong, Niu, and
  Sugiyama]{xia2019anchor}
Xiaobo Xia, Tongliang Liu, Nannan Wang, Bo~Han, Chen Gong, Gang Niu, and
  Masashi Sugiyama.
\newblock Are anchor points really indispensable in label-noise learning?
\newblock In \emph{NeurIPS}, pages 6835--6846, 2019.

\bibitem[Xiao et~al.(2017)Xiao, Rasul, and Vollgraf]{xiao2017fashion}
Han Xiao, Kashif Rasul, and Roland Vollgraf.
\newblock Fashion-mnist: a novel image dataset for benchmarking machine
  learning algorithms.
\newblock \emph{arXiv preprint arXiv:1708.07747}, 2017.

\bibitem[Xiao et~al.(2015)Xiao, Xia, Yang, Huang, and Wang]{xiao2015learning}
Tong Xiao, Tian Xia, Yi~Yang, Chang Huang, and Xiaogang Wang.
\newblock Learning from massive noisy labeled data for image classification.
\newblock In \emph{CVPR}, pages 2691--2699, 2015.

\bibitem[Xu et~al.(2019)Xu, Cao, Kong, and Wang]{xu2019l_dmi}
Yilun Xu, Peng Cao, Yuqing Kong, and Yizhou Wang.
\newblock L\_dmi: A novel information-theoretic loss function for training deep
  nets robust to label noise.
\newblock In \emph{NeurIPS}, pages 6222--6233, 2019.

\bibitem[Yao et~al.(2020{\natexlab{a}})Yao, Yang, Han, Niu, and
  Kwok]{yao2020searching}
Quanming Yao, Hansi Yang, Bo~Han, Gang Niu, and James~T Kwok.
\newblock Searching to exploit memorization effect in learning with noisy
  labels.
\newblock In \emph{ICML}, 2020{\natexlab{a}}.

\bibitem[Yao et~al.(2020{\natexlab{b}})Yao, Liu, Han, Gong, Deng, Niu, and
  Sugiyama]{yao2020dual}
Yu~Yao, Tongliang Liu, Bo~Han, Mingming Gong, Jiankang Deng, Gang Niu, and
  Masashi Sugiyama.
\newblock Dual t: Reducing estimation error for transition matrix in
  label-noise learning.
\newblock In \emph{NeurIPS}, 2020{\natexlab{b}}.

\bibitem[Yao et~al.(2020{\natexlab{c}})Yao, Liu, Han, Gong, Niu, Sugiyama, and
  Tao]{yao2020towards}
Yu~Yao, Tongliang Liu, Bo~Han, Mingming Gong, Gang Niu, Masashi Sugiyama, and
  Dacheng Tao.
\newblock Towards mixture proportion estimation without irreducibility.
\newblock \emph{arXiv preprint arXiv:2002.03673}, 2020{\natexlab{c}}.

\bibitem[Yoo and Choi(2010)]{yoo2010nonnegative}
Jiho Yoo and Seungjin Choi.
\newblock Nonnegative matrix factorization with orthogonality constraints.
\newblock \emph{Management Science}, 58\penalty0 (11):\penalty0 2037--2056,
  2010.

\bibitem[Yu et~al.(2019)Yu, Han, Yao, Niu, Tsang, and Sugiyama]{yu2019does}
Xingrui Yu, Bo~Han, Jiangchao Yao, Gang Niu, Ivor~W Tsang, and Masashi
  Sugiyama.
\newblock How does disagreement benefit co-teaching?
\newblock In \emph{ICML}, 2019.

\bibitem[Yu et~al.(2018{\natexlab{a}})Yu, Liu, Gong, Batmanghelich, and
  Tao]{yu2018efficient}
Xiyu Yu, Tongliang Liu, Mingming Gong, Kayhan Batmanghelich, and Dacheng Tao.
\newblock An efficient and provable approach for mixture proportion estimation
  using linear independence assumption.
\newblock In \emph{CVPR}, pages 4480--4489, 2018{\natexlab{a}}.

\bibitem[Yu et~al.(2018{\natexlab{b}})Yu, Liu, Gong, and Tao]{yu2018learning}
Xiyu Yu, Tongliang Liu, Mingming Gong, and Dacheng Tao.
\newblock Learning with biased complementary labels.
\newblock In \emph{ECCV}, pages 68--83, 2018{\natexlab{b}}.

\bibitem[Yu et~al.(2020)Yu, Liu, Gong, Zhang, Batmanghelich, and
  Tao]{yu2020label}
Xiyu Yu, Tongliang Liu, Mingming Gong, Kun Zhang, Kayhan Batmanghelich, and
  Dacheng Tao.
\newblock Label-noise robust domain adaptation.
\newblock In \emph{ICML}, 2020.

\bibitem[Zhang and Sabuncu(2018)]{zhang2018generalized}
Zhilu Zhang and Mert Sabuncu.
\newblock Generalized cross entropy loss for training deep neural networks with
  noisy labels.
\newblock In \emph{NeurIPS}, pages 8778--8788, 2018.

\end{thebibliography}

\newpage
\appendix
\section{How to learn robust classifiers by exploiting part-dependent transition matrices}
For those who are not familiar with how to use the transition matrix to learn robust classifiers, in this supplementary material, we will provide how to learn robust classifiers by exploiting part-dependent transition matrices. 

We begin by introducing notation. Let $D$ be the distribution of the variables $(X,Y)$, $\bar{D}$ the distribution of the variables $(X,\bar{Y})$. Let $S=\{(\bm{x}_i,y_i)\}_{i=1}^{n}$ be i.i.d. samples drawn from the distribution $D$, $\bar{S}=\{(\bm{x}_i,\bar{y}_i)\}_{i=1}^{n}$  i.i.d. samples drawn from the distribution $\bar{D}$, and $c$ the size of label classes. 

The aim of multi-class classification is to learn a classifier $f$ that can assign labels for given instances. The classifier $f$ is of the following form: $f(\bm{x})=\arg\max_{i\in\{1,2,\ldots,c\}}g_i(\bm{x})$, where $g_i(\bm{x})$ is an estimate of Pr$(Y=i|X=\bm{x})$.
\textit{Expected risk} of employing $f$ is defined as 
\begin{eqnarray}
R(f)=\mathbb{E}_{(X,Y)\sim D}[\ell(f(X),Y)].
\end{eqnarray}
The optimal classifier to learn is the one that minimizes the risk $R(f)$. 
Due to the distribution $D$ is usually unknown, the optimal classifier is approximated by the minimizer of the \textit{empirical risk}:
\begin{eqnarray}
R_n(f)=\frac{1}{n}\sum_{i=1}^{n}\ell(f(\bm{x}_i),y_i).
\end{eqnarray}

Given only the noisy training samples $\{(\bm{x}_i,\bar{y}_i)\}_{i=1}^{n}$, the noisy version of the empirical risk is defined as: 
\begin{eqnarray}
\bar{R}_n(f)=\frac{1}{n}\sum_{i=1}^{n}\ell({f}(\bm{x}_i),\bar{y}_i).
\end{eqnarray}
In the main paper (Section \textcolor{darkblue}{3}), we show how to approximate \textit{instance-dependent} transition matrix by exploiting \textit{part-dependent} transition matrices. For an instance $\bm{x}$,  according to the definition of \textit{instance-dependent} transition matrix, we have that Pr$({\bf \bar{Y}}|X=\bm{x})=T^\top(\bm{x})$Pr$({\bf {Y}}|X=\bm{x})$, we let 
\begin{eqnarray}
\bar{h}(\bm{x})=\arg\max_{i\in\{1,2,\ldots,c\}}(T^\top(\bm{x}){g})_i(\bm{x}).
\label{h}
\end{eqnarray}

The empirical risk of our PTD-F algorithm is defined as: 
\begin{eqnarray}
\bar{R}_n(\bar{h})=\frac{1}{n}\sum_{i=1}^{n}\ell(\bar{h}(\bm{x}_i),\bar{y}_i).
\end{eqnarray}

By employing the \textit{importance reweighting} technique \citep{gretton2009covariate, liu2016classification, xia2019anchor}, the empirical risk of our PTD-R algorithm is defined as:
\begin{align}
\label{eq:em_importance}%
{
\bar{R}_{n}(f,\bar{h})=\frac{1}{n}\sum_{i=1}^{n}\frac{{g}_{\bar{y}_i}(\bm{x}_i)}{\bar{h}_{\bar{y}_i}(\bm{x}_i)}\ell(f(\bm{x}_i),\bar{y}_i)}.
\end{align}
Here, $g_j(\bm{x})$ is an estimate for Pr$(Y=j|\bm{x})$ and  $h_j(\bm{x})$ is an estimate for Pr$(\bar{Y}=j|\bm{x})$. 

When the slack variable $\Delta T$ is introduced to modify the instance-dependent transition matrices, reviewing Eq.~(\ref{h}), we replace $T(\bm{x})$ with $T(\bm{x})+\Delta T$ to get $\bar{h}'(\bm{x})$, i.e., 
\begin{eqnarray}
\bar{h}'(\bm{x})=\arg\max_{i\in\{1,2,\ldots,c\}}(T(\bm{x})+\Delta T)^\top{g})_i(\bm{x}).
\end{eqnarray}

Then the empirical risks of PTD-F-V and PTD-R-V are defined as $\bar{R}_n(\bar{h}')$ and $\bar{R}_n(f,\bar{h}')$, i.e.,
\begin{eqnarray}
\bar{R}_n(\bar{h}')=\frac{1}{n}\sum_{i=1}^{n}\ell(\bar{h}'(\bm{x}_i),\bar{y}_i).
\end{eqnarray}
and
\begin{align}
{
\bar{R}_{n}(f,\bar{h}')=\frac{1}{n}\sum_{i=1}^{n}\frac{{g}_{\bar{y}_i}(\bm{x}_i)}{\bar{h}'_{\bar{y}_i}(\bm{x}_i)}\ell(f(\bm{x}_i),\bar{y}_i)}.
\end{align}

To learn noise robust classifiers under noisy supervision, we minimize the empirical risk of PTD-F, PTD-R, PTD-F-V, and PTD-R-V, respectively.

\section{Instance-dependent Label Noise Generation}
\begin{algorithm}[t]
\setcounter{algorithm}{1}
 {\bfseries Input}: Clean samples $\{(\bm{x}_i, y_i)\}_{i=1}^{n}$; Noise rate $\tau$.
 
	1: Sample instance flip rates $q\in\mathbb{R}^{n}$ from the truncated normal distribution $\mathcal{N}(\tau,0.1^2,[0,1])$;
	
	2: Independently sample $w_1,w_2,\ldots,w_c$ from the standard normal distribution $\mathcal{N}(0,1^2)$;
	
	3: For $i=1,2,\ldots,n$ do
	
    4:\quad $p=\bm{x}_i\times w_{y_i}$;\hfill//generate instance-dependent flip rates
    
    5:\quad  $p_{y_i}=-\infty$;\hfill//control the diagonal entry of the instance-dependent transition matrix
    
    6:\quad  $p=q_i\times softmax(p)$;\hfill//make the sum of the off-diagonal entries of the $y_i$-th row to be $q_i$
    
    7:\quad  $p_{y_i}=1-q_i$;\hfill//set the diagonal entry to be 1-$q_i$
    
    8:\quad  Randomly choose a label from the label space according to the possibilities $p$ as noisy label $\bar{y}_i$;
    
	9: End for.
	
{\bfseries Output}: Noisy samples $\{(\bm{x}_i, \bar{y}_i)\}_{i=1}^{n}$
\caption{Instance-dependent Label Noise Generation}
\label{alg:noise_}
\end{algorithm}

Note that it is more realistic that different instances have different flip rates. Without constraining different instances to have a same flip rate, it is more challenging to model the label noise and train robust classifiers. In Step~1, in order to control the global flip rate as $\tau$ but without constraining all of the instances to have a same flip rate, we sample their flip rates from a truncated normal distribution $\mathcal{N}(\tau,0.1^2,[0,1])$.  Specifically, this distribution limits the flip rates of instances in the range $[0,1]$. Their mean and standard deviation are equal to the mean $\tau$ and the standard deviation 0.1 of the selected truncated normal distribution respectively. 

In Step~2, we sample parameters $w_1,w_2,\ldots,w_c$ from the standard normal distribution for generating \textit{instance-dependent} label noise. The dimensionality of each parameter is $d\times c$, where $d$ denotes the dimensionality of the instance. Learning these parameters is critical to model \textit{instance-dependent} label noise. However, it is hard to identify these parameters without any assumption. 

Note that an instance with clean label $y$ will be flipped only according to the $y$-th row of the transition matrix. Thus, in Steps~4 to 7, we only use the $y_i$-th row of the \textit{instance-dependent} transition matrix for the instance $\bm{x}_i$. Specifically, Steps~5 and 7 are to ensure the diagonal entry of the $y_i$-th row is 1- $q_i$. Step~6 is to ensure that the sum of the off-diagonal entries is $q_i$.

\begin{table}[!htbp]
	\centering
	\caption{Means and standard deviations (percentage) of classification accuracy on \textit{MNIST} with different label noise levels. }
	\label{tab:accu1}
	{
		\begin{tabular}{cccccc}
           	\toprule
			 ~& IDN-10\%      & IDN-20\%      & IDN-30\%    & IDN-40\%  & IDN-50\%         \\ \midrule
			 CE & 98.24$\pm$0.07 & 98.21$\pm$0.06 & 96.78$\pm$0.12 & 93.76$\pm$0.18 & 79.69$\pm$4.35\\
             Decoupling  & 96.63$\pm$0.12 & 96.62$\pm$0.22 & 92.73$\pm$0.36 & 90.34$\pm$0.33 & 80.56$\pm$2.67 \\ 
		     MentorNet & 97.45$\pm$0.11 & 97.21$\pm$0.13 & 92.88$\pm$0.31 & 88.23$\pm$1.65 & 80.02$\pm$1.71 \\
			 Co-teaching & 97.56$\pm$0.12 & 97.32$\pm$0.15 & 94.81$\pm$0.24  & 92.45$\pm$0.59 & 83.30$\pm$1.37 \\
			 Co-teaching+ & 98.32$\pm$0.07 & 98.07$\pm$0.12 & 96.70$\pm$0.35  & 94.37$\pm$0.48 & 82.97$\pm$1.11 \\
			 Joint & 98.53$\pm$0.06 & 98.17$\pm$0.14 & 96.51$\pm$0.17  & 93.07$\pm$0.62 & 83.72$\pm$3.22 \\
			 DMI & 98.63$\pm$0.04 & 98.40$\pm$0.11 & 97.75$\pm$0.21  & 96.45$\pm$0.23 & 87.52$\pm$1.03 \\
			 Forward & 97.23$\pm$0.15 & 96.87$\pm$0.15 & 95.01$\pm$0.27  & 90.30$\pm$0.61 & 77.42$\pm$3.28 \\
			 Reweight & 98.21$\pm$0.07 & 97.99$\pm$0.13 & 96.96$\pm$0.14  & 94.55$\pm$0.67 & 80.87$\pm$4.14 \\
			 T-Revision & 98.49$\pm$0.06 & 98.39$\pm$0.09 & 97.55$\pm$0.14  & 96.50$\pm$0.31 & 84.71$\pm$3.47 \\
			 \midrule
			 PTD-F& 98.55$\pm$0.05 & 97.92$\pm$0.27 & 97.34$\pm$0.11  & 94.67$\pm$0.83 & 84.01$\pm$2.11 \\
			 PTD-R & 98.22$\pm$0.10 & 98.12$\pm$0.17 & 97.06$\pm$0.13  & 94.75$\pm$0.54 & 82.72$\pm$2.04 \\
			 PTD-F-V & \textbf{98.71$\pm$0.05} & \textbf{98.46$\pm$0.11} & 97.77$\pm$0.09  & 96.07$\pm$0.45 & 88.55$\pm$1.96 \\
			 PTD-R-V & 98.66$\pm$0.03 & 98.43$\pm$0.15 & \textbf{97.81$\pm$0.23}  & \textbf{96.73$\pm$0.20} & \textbf{88.67$\pm$1.25} \\
             \bottomrule
		\end{tabular}
	}
\end{table}

\section{Experiments complementary on synthetic noisy datasets}
In the main paper (Section \textcolor{darkblue}{4}), we present the experimental results on four synthetic noisy datasets, i.e., \textit{F-MNIST}, \textit{SVHN}, \textit{CIFAR-10}, and \textit{NEWS}. In this supplementary material, we provide the experimental results on another synthetic noisy dataset \textit{MNIST}. \textit{MNIST} contains 60,000 training images and 10,000 test images with 10 classes. We use a LeNet-5 network for it. The detailed experimental results are shown in Table \ref{tab:accu1}. The classification performance shows that our proposed method is more robust than the baseline methods when coping with \textit{instance-dependent} label noise.

\section{The details of significance tests}
We exploit significance tests to show whether all experimental results are statistically significant. The \textit{p}-values are obtained with two independent samples t-test \cite{sedgwick2010independent}. Note that small \textit{p}-values reflect the performance of the proposed method is significantly better than the performance of the baselines. The proposed method PTD-R-V achieves the best classification performance in almost all cases. We thus conduct significance tests to compare the baselines with PTD-R-V. The results of significance tests are presented in Table \ref{tab:sign}. We can see that almost all results are statistically significant.

\begin{table}[!t]
	\centering
	\setlength\tabcolsep{2pt}
	\caption{The results of significant tests (\textit{p}-value) on five synthetic noisy datasets with different noise levels.}\label{tab:sign}
	\begin{tabular}{p{2cm}p{2.5cm}p{1.7cm}<{\centering}p{1.7cm}<{\centering}p{1.7cm}<{\centering}p{1.7cm}<{\centering}p{1.7cm}<{\centering}}
		\Xhline{3\arrayrulewidth}
		Dataset & Method &  IDN-10\% & IDN-20\% & IDN-30\% & IDN-40\% & IDN-50\%\\
		\hline
		\multirow{10}{*}{\textit{MNIST}}& CE & 0.0000 & 0.0187 & 0.0000 & 0.0000 & 0.0152  \\ 
		& Decoupling  & 0.0000 & 0.0000 & 0.0000 & 0.0000 & 0.0002  \\ 
		& MentorNet  & 0.0000 & 0.0000 & 0.0000 & 0.0000 & 0.0004 \\ 
		& Co-teaching  & 0.0000 & 0.0000 & 0.0000 & 0.0000 & 0.0001  \\
		& Co-teaching+  & 0.0000 & 0.0005 & 0.0009 & 0.0000 & 0.0000  \\
		& Joint & 0.0036 & 0.0202  & 0.0000 & 0.0000 & 0.0112 \\ 
		& DMI & 0.0016 & 0.4489  & 0.0157 & 0.1732 & 0.4395 \\ 
		& Forward  & 0.0000 & 0.0000 & 0.0000 & 0.0000 & 0.0000 \\
		& Reweight  & 0.0000 & 0.0044 & 0.0001 & 0.0011 & 0.0025 \\
		& T-Revision & 0.0050 & 0.8045 & 0.0058 & 0.2130 & 0.0695 \\
		\hline	
		\hline 
		\multirow{10}{*}{\textit{F-MNIST}}& CE & 0.0000 & 0.0001 & 0.0000 & 0.0000 & 0.0000   \\ 
		& Decoupling  & 0.0000 & 0.0000 & 0.0000 & 0.0000 & 0.0000  \\ 
		& MentorNet  & 0.0000 & 0.0000 & 0.0003 & 0.0001 & 0.0000 \\ 
		& Co-teaching  & 0.0000 & 0.0000 & 0.0000 & 0.0173 & 0.0000  \\
		& Co-teaching+  & 0.0737 & 0.0023 & 0.0030 & 0.4358 & 0.0000  \\
		& Joint & 0.0000 & 0.0000  & 0.0000 & 0.0000 & 0.0000  \\ 
		& DMI & 0.4281 & 0.0068 & 0.0000 & 0.0000 & 0.0000 \\ 
		& Forward  & 0.0001 & 0.0002 & 0.0000 & 0.0000 & 0.0000 \\
		& Reweight  & 0.0000 & 0.0041 & 0.0000 & 0.0001 & 0.0001 \\
		& T-Revision & 0.9522 & 0.1335 & 0.1626 & 0.0931 & 0.0002 \\
		\hline	
		\hline
		\multirow{10}{*}{\textit{SVHN}}& CE & 0.0000 & 0.0000 & 0.0000 & 0.0000 & 0.0012 \\ 
		& Decoupling  & 0.0000 & 0.0000 & 0.0000 & 0.0005 & 0.0000  \\ 
		& MentorNet  & 0.0000 & 0.0000 & 0.0001 & 0.0000 & 0.0000 \\ 
		& Co-teaching  & 0.0000 & 0.0000 & 0.0001 & 0.0000 & 0.0000  \\
		& Co-teaching+  & 0.0001 & 0.0000 & 0.0000 & 0.0000 & 0.0005  \\
		& Joint & 0.0000 & 0.0000  & 0.0000 & 0.0000 & 0.0001  \\ 
		& DMI & 0.0068 & 0.0385 & 0.6901 & 0.0000 & 0.0002 \\ 
		& Forward  & 0.0000 & 0.0000 & 0.0002 & 0.0000 & 0.0000 \\
		& Reweight  & 0.0001 & 0.0139 & 0.0031 & 0.0018 & 0.0002 \\
		& T-Revision & 0.2258 & 0.3116 & 0.5436 & 0.0471 & 0.0228 \\
		\hline		
		\hline 
	    \multirow{10}{*}{\textit{CIFAR-10}}& CE & 0.0000 & 0.0000 & 0.0000 & 0.0000 & 0.0000   \\ 
		& Decoupling  & 0.0001 & 0.0000 & 0.0000 & 0.0000 & 0.0000  \\ 
		& MentorNet  & 0.0000 & 0.0000 & 0.0000 & 0.0000 & 0.0000 \\ 
		& Co-teaching  & 0.0000 & 0.0000 & 0.0000 & 0.0000 & 0.0000  \\
		& Co-teaching+  & 0.0002 & 0.0001 & 0.0000 & 0.0001 & 0.0000  \\
		& Joint & 0.0000 & 0.0000  & 0.0083 & 0.0704 & 0.0638  \\ 
		& DMI & 0.0000 & 0.0000 & 0.0000 & 0.0000 & 0.0000 \\ 
		& Forward  & 0.0000 & 0.0000 & 0.0000 & 0.0000 & 0.0000 \\
		& Reweight  & 0.0000 & 0.0000 & 0.0000 & 0.0000 & 0.0000 \\
		& T-Revision & 0.0000 & 0.0000 & 0.0000 & 0.0000 & 0.0013 \\
		\hline		
		\hline 
		\multirow{10}{*}{\textit{NEWS}}& CE & 0.0000 & 0.0000 & 0.0000 & 0.0000 & 0.0002   \\ 
		& Decoupling  & 0.0000 & 0.0000 & 0.0000 & 0.0000 & 0.0000  \\ 
		& MentorNet  & 0.0000 & 0.0001 & 0.0000 & 0.0000 & 0.0000 \\ 
		& Co-teaching  & 0.0000 & 0.0027 & 0.0125 & 0.0000 & 0.0000  \\
		& Co-teaching+  & 0.0000 & 0.0001 & 0.0000 & 0.0000 & 0.0000  \\
		& Joint & 0.0000 & 0.0000 & 0.0001 & 0.0025 & 0.0008  \\ 
		& DMI & 0.0004 & 0.0000 & 0.0004 & 0.0032 & 0.0001 \\ 
		& Forward  & 0.0000 & 0.0000 & 0.0000 & 0.0000 & 0.0000 \\
		& Reweight  & 0.0000 & 0.0001 & 0.0008 & 0.0021 & 0.0108 \\
		& T-Revision & 0.0010 & 0.0006 & 0.0040 & 0.0052 & 0.0285 \\
		\Xhline{3\arrayrulewidth}
	\end{tabular}
\end{table}

\section{The experimental results of ablation study}
In Section \textcolor{darkblue}{4.2}, we have shown that our proposed method is insensitive to the number of parts. Due the space limit, we only provide the illustration by exploiting the figures. In this supplementary material, more detailed results including means and standard deviations of approximation error and classification accuracy about the ablation study are shown in Table \ref{tab:approximation} and Table \ref{tab:acc6}.

\begin{table}[!htbp]
	\centering
	\small
	\caption{Means and standard deviations of approximation error on \textit{CIFAR-10} with 50\% label noise level.}
	\label{tab:approximation}
	{
		\begin{tabular}{cccccc}
           	\toprule
			 ~& Class-dependent & T-Revision & PTD & PTD-F-V &  PTD-R-V     \\ \midrule
             $r$=10  & 0.945$\pm$0.051 & 0.922$\pm$0.037 & 0.840$\pm$0.030 & 0.815$\pm$0.011 & 0.811$\pm$0.020 \\ 
		     $r$=11 & 0.945$\pm$0.051 & 0.922$\pm$0.037 & 0.841$\pm$0.022 & 0.802$\pm$0.010 & 0.815$\pm$0.011 \\
			 $r$=12 & 0.945$\pm$0.051 & 0.922$\pm$0.037 & 0.831$\pm$0.015  & 0.806$\pm$0.014 & 0.812$\pm$0.014 \\
			 $r$=13 & 0.945$\pm$0.051 & 0.922$\pm$0.037 & 0.814$\pm$0.024  & 0.790$\pm$0.019 & 0.791$\pm$0.017 \\
			 $r$=14 & 0.945$\pm$0.051 & 0.922$\pm$0.037 & 0.821$\pm$0.040 & 0.792$\pm$0.022 & 0.791$\pm$0.016\\
			 $r$=15 & 0.945$\pm$0.051 & 0.922$\pm$0.037 & 0.829$\pm$0.034  & 0.812$\pm$0.017 & 0.802$\pm$0.025 \\
			 $r$=16 & 0.945$\pm$0.051 & 0.922$\pm$0.037 & 0.831$\pm$0.029  & 0.800$\pm$0.018 & 0.800$\pm$0.020 \\
			 $r$=17 & 0.945$\pm$0.051 & 0.922$\pm$0.037 & 0.819$\pm$0.012  & 0.800$\pm$0.011 & 0.792$\pm$0.013 \\
			 $r$=18 & 0.945$\pm$0.051 & 0.922$\pm$0.037 & 0.829$\pm$0.011  & 0.798$\pm$0.012 & 0.794$\pm$0.017 \\
			 $r$=19 & 0.945$\pm$0.051 & 0.922$\pm$0.037 & 0.827$\pm$0.017  & 0.799$\pm$0.013 & 0.795$\pm$0.018 \\
			 $r$=20 & 0.945$\pm$0.051 & 0.922$\pm$0.037 & 0.832$\pm$0.025 & 0.805$\pm$0.021 & 0.800$\pm$0.015\\
             \bottomrule
		\end{tabular}
	}
\end{table}

\begin{table}[!htbp]
	\centering
	\caption{Means and standard deviations (percentage) of classifation accuracy on \textit{CIFAR-10} with 50\% label noise level.}
	\label{tab:acc6}
	{
		\begin{tabular}{ccccc}
           	\toprule
			 ~& PTD-F & PTD-R & PTD-F-V & PTD-R-V      \\ \midrule
             $r$=10 & 46.84$\pm$2.34 & 49.02$\pm$2.55 & 48.84$\pm$2.74 & 53.78$\pm$2.77  \\ 
		     $r$=11 & 47.22$\pm$1.77 & 49.11$\pm$1.98 & 48.64$\pm$1.58 & 53.72$\pm$2.63   \\
			 $r$=12 & 47.01$\pm$2.65 & 48.75$\pm$1.95 & 48.62$\pm$3.05 & 53.52$\pm$1.99   \\
			 $r$=13 & 47.05$\pm$1.87 & 48.99$\pm$2.67 & 48.63$\pm$1.42 & 53.33$\pm$1.96   \\
			 $r$=14 & 47.01$\pm$1.65 & 49.12$\pm$3.02 & 48.77$\pm$1.46 & 53.72$\pm$2.13  \\
			 $r$=15 & 46.88$\pm$1.29 & 49.14$\pm$1.89 & 48.65$\pm$1.01 & 53.90$\pm$1.67   \\
			 $r$=16 & 47.19$\pm$1.49 & 49.03$\pm$1.78 & 48.59$\pm$2.03 & 53.98$\pm$1.95   \\
			 $r$=17 & 47.01$\pm$1.36 & 49.02$\pm$2.06 & 48.62$\pm$1.62 & 54.01$\pm$1.72   \\
			 $r$=18 & 47.09$\pm$1.45 & 48.89$\pm$2.51 & 48.58$\pm$1.03 & 53.69$\pm$2.31   \\
			 $r$=19 & 47.39$\pm$1.48 & 49.09$\pm$2.58 & 48.79$\pm$1.01 & 53.75$\pm$2.77   \\
			 $r$=20 & 46.88$\pm$1.25 & 49.07$\pm$2.56 & 48.76$\pm$1.75 & 53.98$\pm$2.34 \\
             \bottomrule
		\end{tabular}
	}
\end{table}

\section{Visualization of parts}
\begin{figure}[!tp]
\centering
\includegraphics[width=12.0cm,height=7.0cm]{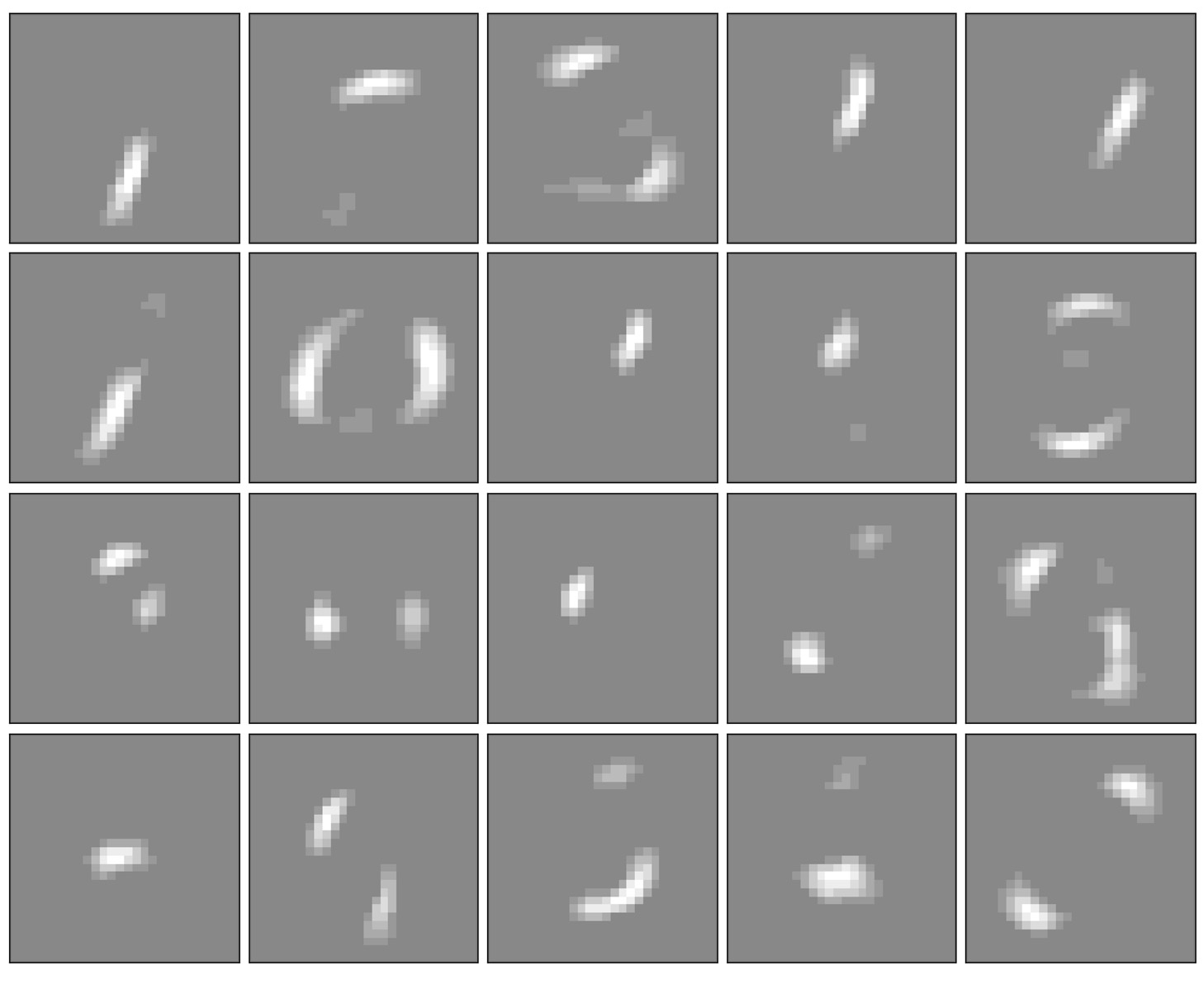}
\caption{Visualization of parts for \textit{MNIST}.}
\label{fig:nmf_mnist}
\end{figure}

\begin{figure}[!tp]
\centering
\includegraphics[width=12.0cm,height=7.0cm]{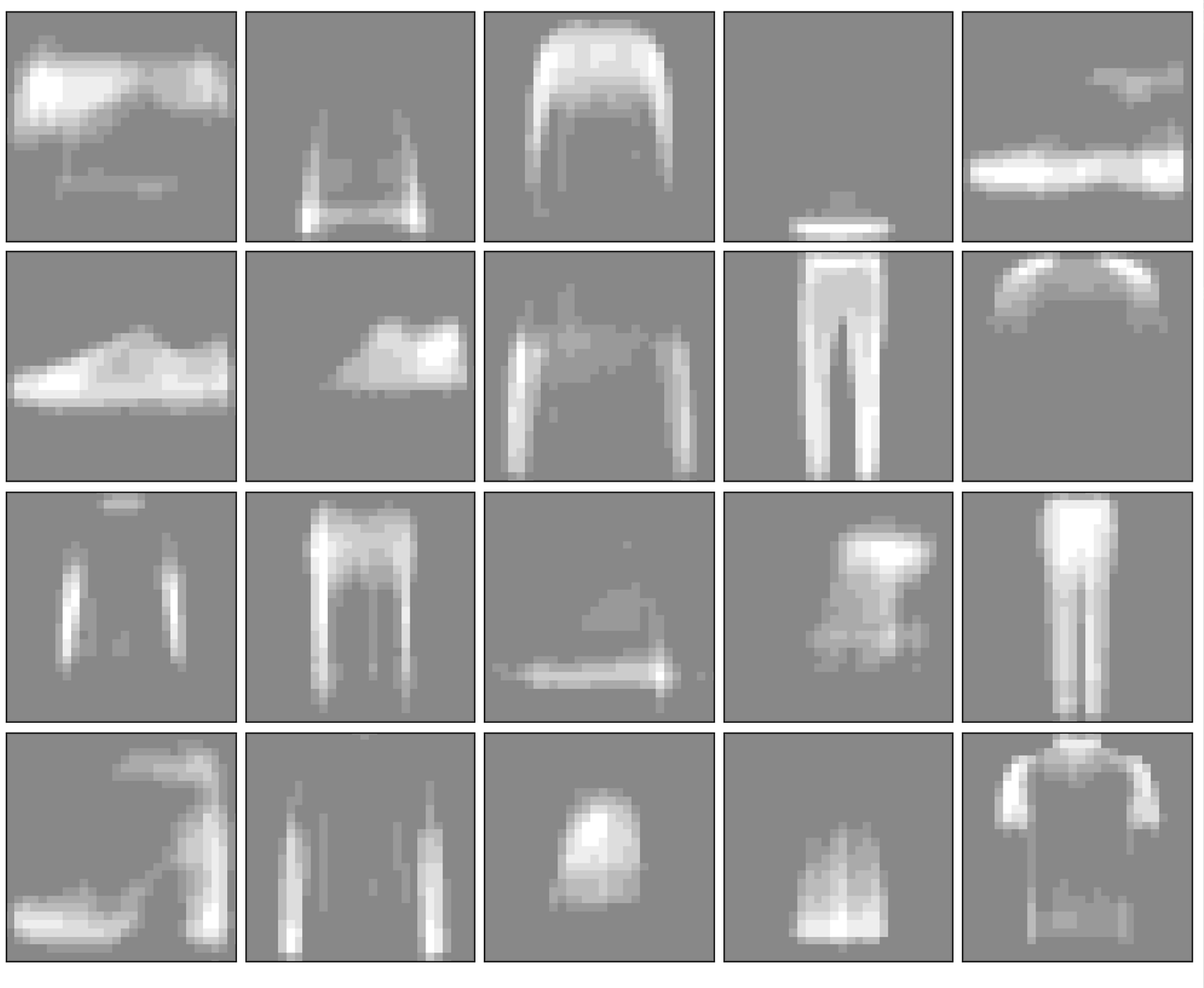}
\caption{Visualization of parts for \textit{F-MNIST}.}
\label{fig:nmf_fmnist}
\end{figure}

Note that to make use of the power of deep learning, in the main paper, the data matrix used for factorization consists of deep representations extracted by a deep network. We learn parts and parts-based representations (new representations) by applying NMF to this data matrix. Although deep representations contain semantic information, it is not easy to visualize these parts obtained from the deep representations directly. We propose to approximate and visualize the parts of the deep representations by studying their corresponding parts of the original observations. Intuitively, let the NMF of the deep representations and the original observations to have the same parts-based representations, the obtained parts from the two factorizations should be corresponding to each other. The obtained parts for \textit{MNIST} and \textit{F-MNIST} are presented in Figure \ref{fig:nmf_mnist} and Figure \ref{fig:nmf_fmnist}.
Note that the datasets, i.e., \textit{SVHN}, \textit{CIFAR-10}, and \textit{Clothing1M}, are also used to verify the effectiveness of the proposed method. However, the instances in these datasets contain three channels (i.e., RGB channels). It is hard to properly visualize the parts of the deep representations by finding their corresponding parts of the original observations. 

\end{document}